\documentclass{article}
\usepackage{ijcai25}

\usepackage{times}
\usepackage{soul}
\usepackage{url}
\usepackage[hidelinks]{hyperref}
\usepackage[utf8]{inputenc}
\usepackage[small]{caption}
\usepackage{graphicx}
\usepackage{amsmath}
\usepackage{amsthm}
\usepackage{booktabs}
\usepackage{algorithm}
\usepackage[switch]{lineno}

\usepackage[toc,page]{appendix}
\usepackage{multirow}
\usepackage{hyperref}
\usepackage{amssymb,amsfonts}
\usepackage{algpseudocode}
\usepackage{amssymb}
\usepackage{amsmath}
\usepackage{mathtools}
\usepackage[normalem]{ulem}
\usepackage[table]{xcolor} 

\theoremstyle{definition}

\usepackage[colorinlistoftodos]{todonotes}

\theoremstyle{remark} 
\newtheorem{remark}{Remark} 
\definecolor{mmbcolor}{RGB}{0, 0, 255} 

\definecolor{hancolor}{RGB}{0,153,0}
\definecolor{bincolor}{RGB}{138,43,226}

\title{Non-collective Calibrating Strategy for Time Series Forecasting}

\author{
    Author Name
    \affiliations
    Affiliation
    \emails
    email@example.com
}

\author{
Bin Wang$^{1 \ast}$ \and
Yongqi Han$^{1 \ast}$ \and
Minbo Ma$^2$\and
Tianrui Li$^2$\and
Junbo Zhang$^{3,4,5}$\and \\
Feng Hong$^{1 \dagger}$ \and
Yanwei Yu$^{1 \dagger}$ 
\affiliations
$^1$Ocean University of China
$^2$Southwest Jiaotong University
$^3$JD Intelligent Cities Research
$^4$JD iCity, JD Technology, China
$^5$Beijing Key Laboratory of Traffic Data Mining and Embodied Intelligence
\emails
wangbin9545@ouc.edu.cn,
hanyuki23@stu.ouc.edu.cn,
minbo46.ma@gmail.com,
msjunbozhang@outlook.com,
trli@swjtu.edu.cn
\{hongfeng,yuyanwei\}@ouc.edu.cn,
}
\begin{document}

\maketitle

\begin{abstract}
Deep learning-based approaches have demonstrated significant advancements in time series forecasting. Despite these ongoing developments, the complex dynamics of time series make it challenging to establish the rule of thumb for designing the golden model architecture. In this study, we argue that refining existing advanced models through a universal calibrating strategy can deliver substantial benefits with minimal resource costs, as opposed to elaborating and training a new model from scratch. We first identify a multi-target learning conflict in the calibrating process, which arises when optimizing variables across time steps, leading to the underutilization of the model's learning capabilities. To address this issue, we propose an innovative calibrating strategy called Socket+Plug (SoP). This approach retains an exclusive optimizer and early-stopping monitor for each predicted target within each Plug while keeping the fully trained Socket backbone frozen. The model-agnostic nature of SoP allows it to directly calibrate the performance of any trained deep forecasting models, regardless of their specific architectures. Extensive experiments on various time series benchmarks and a spatio-temporal meteorological ERA5 dataset demonstrate the effectiveness of SoP, achieving up to a 22\% improvement even when employing a simple MLP as the Plug (highlighted in \autoref{fig:all}). Code is available at \textit{\url{https://github.com/hanyuki23/SoP}.}
\end{abstract}

\renewcommand{\thefootnote}{\relax}
\footnotetext[1]{$^\ast$ Bin Wang and Yongqi Han are co-first authors with equal contributions.}
\footnotetext[1]{$^\dagger$ Feng Hong and Yanwei Yu are corresponding authors.}
\renewcommand{\thefootnote}{\arabic{footnote}}

\begin{figure}[ht]
  \centering
  \includegraphics[width=0.95\linewidth]{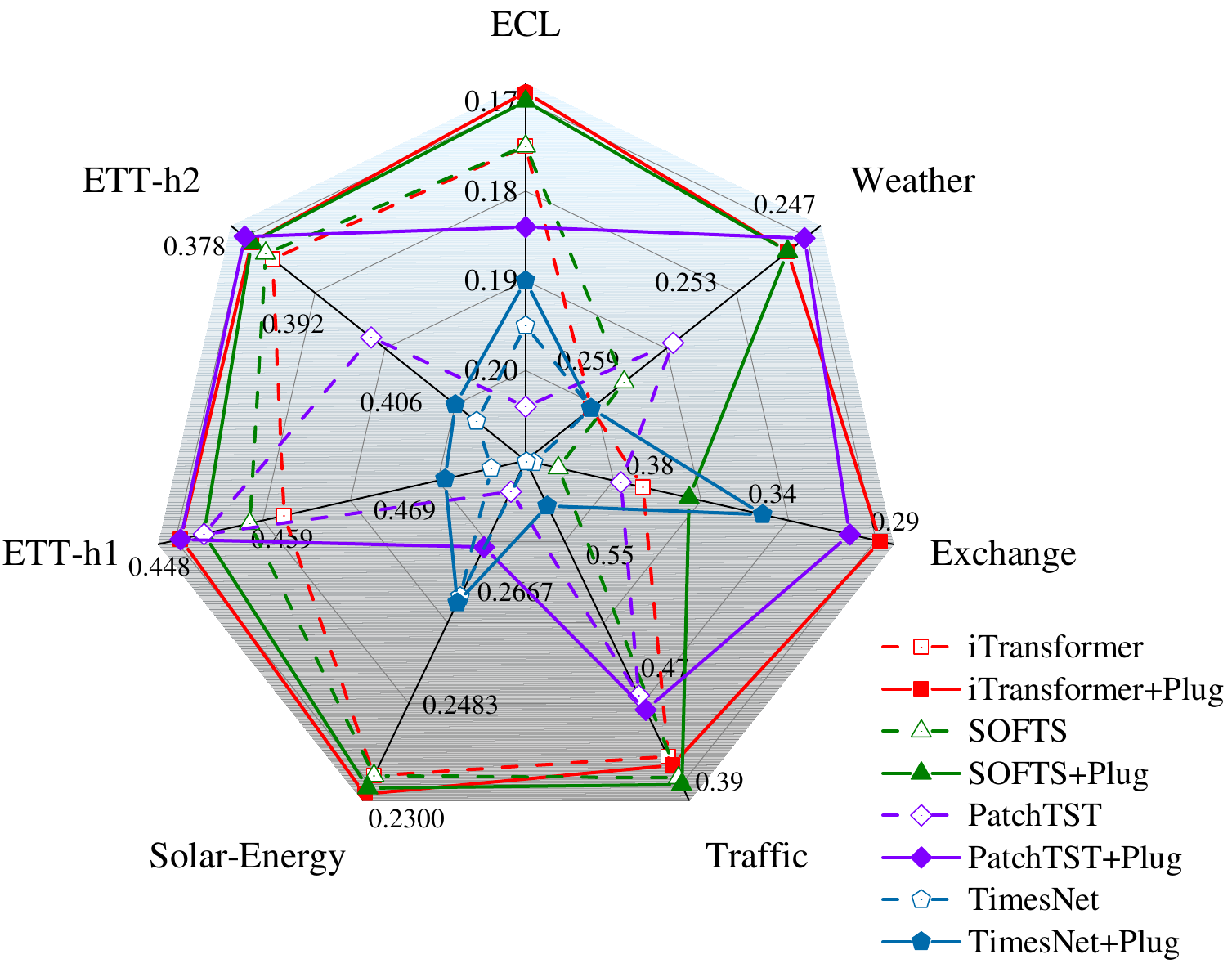}
  \caption{
  Performance on popular benchmarks is reported when employing the proposed universal calibrating strategy, denoted as \textit{Model+Plug} and represented by the solid line. 
  }
  \label{fig:all}
\end{figure}

\section{Introduction}
Time series forecasting remains challenging due to the inherent diversity and complexity of the data, which are influenced by factors such as seasonal fluctuations, trends, and domain-specific patterns~\cite{deng2024disentangling}.  To investigate the most effective deep learning architecture, researchers have endeavored from {convolutional neural networks (CNNs)~\cite{2017Conditional} and recurrent neural networks (RNNs)~\cite{LSTNET18} to graph neural networks (GNNs)~\cite{wen2023diffstg}} and Transformers~\cite{zhang2024probts}. However, empirical debates persist regarding the performance of these models in time series forecasting. For instance, some studies have revisited the argument that a properly designed simple multi-layer perceptron (MLP) can outperform advanced Transformer-based models~\cite{Zeng_Chen_Zhang_Xu_2023_dlnear,lu2018structural}. {With SOTA models emerging almost daily, disagreements over performance can incur significant inefficiencies for decision-makers during model selection. Consider a scenario where a trained model is already performing effectively in a production environment, yet numerous novel SOTA models continue to emerge. Determining whether any of these new models has the potential to surpass the deployed model typically requires extensive trial-and-error experimentation. This raises a critical question: rather than exhaustively testing each new SOTA model, could the existing trained model be enhanced at minimal cost?}

\begin{figure*}[!tbh]
\centering
\includegraphics[width=0.95\linewidth]{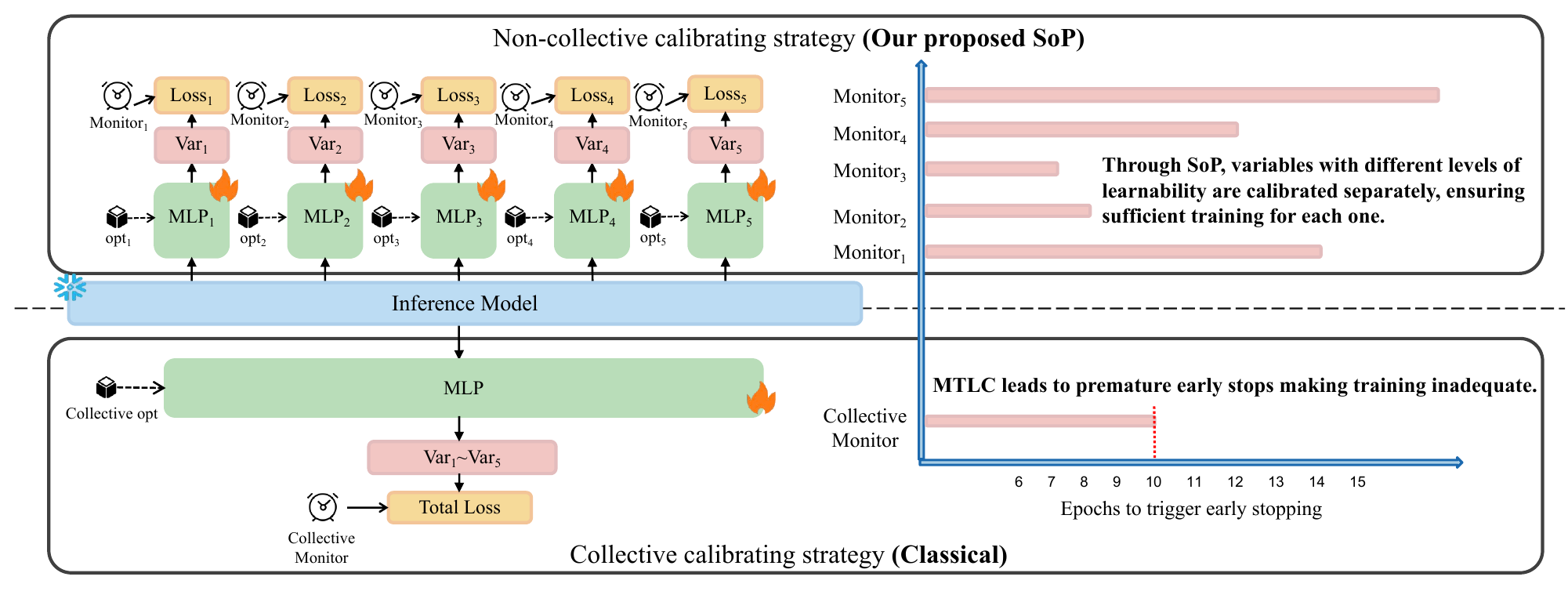}
    \caption{This banner highlights the distinct effects between SoP and the classical collective calibrating strategy. The MTLC phenomenon often causes premature early stopping, resulting in insufficient training. In contrast, non-collective SoP  allows target variables with different levels of learnability to be calibrated independently, ensuring adequate and effective training for each. 
}
\label{fig：haibao}
\end{figure*}

In this study, we instead draw attention to harness trained time series forecasting methods rather than designing a new model from scratch. An intuitive solution involves calibrating time series predictions using prior knowledge, such as rule-based methods that apply exponential smoothing and filtering to reduce noise and constraint bounds. Despite their simplicity, such approaches demand rigorous domain knowledge, as improper handling can degrade performance. 
Motivated by the effectiveness of learning-based classification calibration \cite{2000Probabilistic}, we would explore its potential in time series forecasting models.

Classical calibrating typically involves keeping the fully trained model frozen while post-processing only the outputs of the forecasting models~\cite{KuleshovFE18}. Generally, this process calibrates the predictive bias across all target variables and prediction horizons simultaneously, a strategy we term as \textit{collective calibrating} in this study. In contrast, when the calibrating process is applied separately to grouped or individual variables or prediction horizons, we introduce the term \textit{non-collective calibrating}. To investigate the effect of these two strategies, we started by conducting a preliminary experiment, with the key discovery illustrated in~\autoref{fig：haibao} and described as follows. 
Employing a classical collective calibrating strategy — where an additional MLP to be finetuned (illustrated in the bottom part of~\autoref{fig：haibao}) is attached to the existing trained inference model (represent by the blue rectangle)— we observed premature triggering  (e.g., around 10 epochs) of the early-stopping monitor. Conversely, when the single MLP was split equally by the number of neurons into five smaller, independent MLPs denoted as MLP$_1$ to MLP$_5$, each dedicated to be finetuned for a single target variable and equipped with its own optimizer and early-stopping monitor (illustrated in the upper part of~\autoref{fig：haibao}), we observed significant variation in the early-stopping epochs across monitors. While some stopped in fewer than 10 epochs, others required substantially more epochs to achieve optimal performance on validation data. Our heuristic hypothesis is that, due to varying levels of learnability across different targets, optimizing all variables simultaneously under the same objective loss function may fail to fully capture the unique distribution of each target variable {— a phenomenon we refer to as multi-target learning conflict (MTLC)}. 

To test the above hypothesis, we have conducted rigorous ablation study to compare the forecasting performance of the two strategies. 
Our findings support that the non-collective calibrating strategy outperforms its collective counterpart regarding generalization performance (referred to Table~\ref{tab:my-table5}). 
We metaphorically term this approach the \textit{Socket+Plug} strategy (SoP), where the well-trained inference model serves as a \textit{Socket}, providing the fundamental shared predictive ability, while each {additional module (e.g., MLP)} as a \textit{Plug}, tailored to its specific calibrating target. Extensive experiments further demonstrate that SoP can generally enhance the performance when existing deep forecasting models are taken as the Socket, as highlighted in \autoref{fig:all}. We further investigated two specific forms of SoP: variable-wise SoP and step-wise SoP. Experimental results indicate that, under the same model architectures, both step-wise and variable-wise SoP exhibit similarly strong performance. It is worth noting that SoP leverages off-the-shelf Socket outputs without requiring any modifications to the Socket, making it easily combined with existing SOTA deep forecasting methods. Moreover, we applied SoP to the spatio-temporal meteorological forecasting task, utilizing Unet as the Socket. The experimental results validate its effectiveness, highlighting its potential for integration with advanced weather forecasting systems, such as PanGu~\cite{bi2022pangu} and FengWu~\cite{chen2023fuxi}, thereby underscoring its applicability to real-world weather forecasting tasks. In summary, the key contributions of this study are summarized as follows: 
\begin{itemize}
\item {To our best knowledge, we are the first to identify and emphasize the detrimental impact of the MTLC phenomenon in deep forecasting, which limits the learning potential of time series models. Specifically, we demonstrate that the traditional early-stopping mechanism in model training could fail to fully exploit the capabilities of these models.}

\item We propose SoP, a universal and model-agnostic calibrating strategy, developed to calibrate the performance of fully trained deep forecasting models with minimal design costs. {By employing a dedicated optimizer and early-stopping monitor for each predicted target, SoP effectively mitigates the MTLC issue.}

\item We derive two specific forms of SoP, referred to as target-wise SoP, which include variable-wise SoP and step-wise SoP. Both demonstrate significantly better performance than classical collective calibrating and can be used as quick-start versions of SoP, eliminating the need to determine the $Counts$ hyperparameter for the Plug.


\item We conduct extensive experiments on seven time series benchmark datasets and the spatio-temporal ERA5 meteorological dataset, whose results demonstrate the effectiveness of SoP, achieving up to a 22\% improvement in forecasting performance. This showcases its potential for advancing time series and spatio-temporal forecasting tasks.
\end{itemize}


\section{Preliminaries}
\subsection{Definitions}
\textbf{Definition 1} (Time Series Forecasting).  
Given a historical observation series $ X = \{ x_{1}, x_{2}, \dots, x_{T} \} \in \mathbb{R}^{N \times T}$, which represents past \( T \) time steps and \( N \) variables, let \( X_{:,t} \) denote the values of all variables at time step \( t \), and \( X_{n,:} \) denote the values across all past time steps for variable \( n \). Time series forecasting aims to learn a model $f(\cdot)$ that predicts the future values \( \bar{Y} = f(X) =  \{ x_{T+1}, x_{T+2}, \dots, x_{T+S} \} \in \mathbb{R}^{N \times S} \) over the future \( S \) time steps, a.k.a., \( S \) prediction horizons.

\noindent \textbf{Definition 2} (Collective \& Non-collective calibrating). Traditional calibrating typically involves keeping the inference model frozen while re-training part of model layers collectively for \textit{all target variables across all prediction horizons}. We define this type of calibrating as \textit{collective calibrating}. In contrast, when the calibrating process is applied separately to grouped or even individual variables or prediction horizons, we denote this as \textit{non-collective calibrating}. With these definitions, our proposed SoP strategy falls under non-collective calibrating. 

\noindent \textbf{Definition 3} (Optimized Plug \& Plug Counts).
When implementing SoP, one can decide how many variables or prediction horizons are optimized together as a group, each with its own optimizer and early-stopping monitor. We refer to each such group of variables or horizons as an \textit{Optimized Plug} and the number of total optimized plugs is denoted as the \textit{Plug Counts}. Specifically, for a prediction target \( Y \in \mathbb{R}^{N \times S} \): If \( n \) variables along the \( N \)-dimension form an optimized plug to predict \( Y_{plug} \in \mathbb{R}^{n \times S} \), SoP creates the plug counts as \( M = \frac{N}{n} \). If \( s \) horizons along the \( S \)-dimension form an optimized plug to predict  \( Y_{plug} \in \mathbb{R}^{N \times s} \), SoP creates the plug counts as \( M = \frac{S}{s} \). Two special remarks are derived as:

\begin{remark}
There are two types of \textit{target-wise SoP}: when \( n = 1 \), \( M = N \), it is referred to as the \textit{variable-wise SoP}; and when \( s = 1 \), \( M = S \), it is referred to as the \textit{step-wise SoP}.
\end{remark}

\begin{remark}
When \( n = N \) and \( s = S \), \( M = 1 \), in which case SoP reduces to the classical collective calibrating strategy.
\end{remark}

\section{Methodology}

To clarify the data flow in the model operation based on SoP, this section first introduces the model inference process given the input, followed by an explanation of the calibrating process of SoP. {Without loss of generality and for clarity, we use a variable-wise SoP (where $M = N$, as described in \textit{Remark} 1.) to illustrate the process.}

\begin{figure*}[!htbp]
\centering
\includegraphics[width=0.8\linewidth]{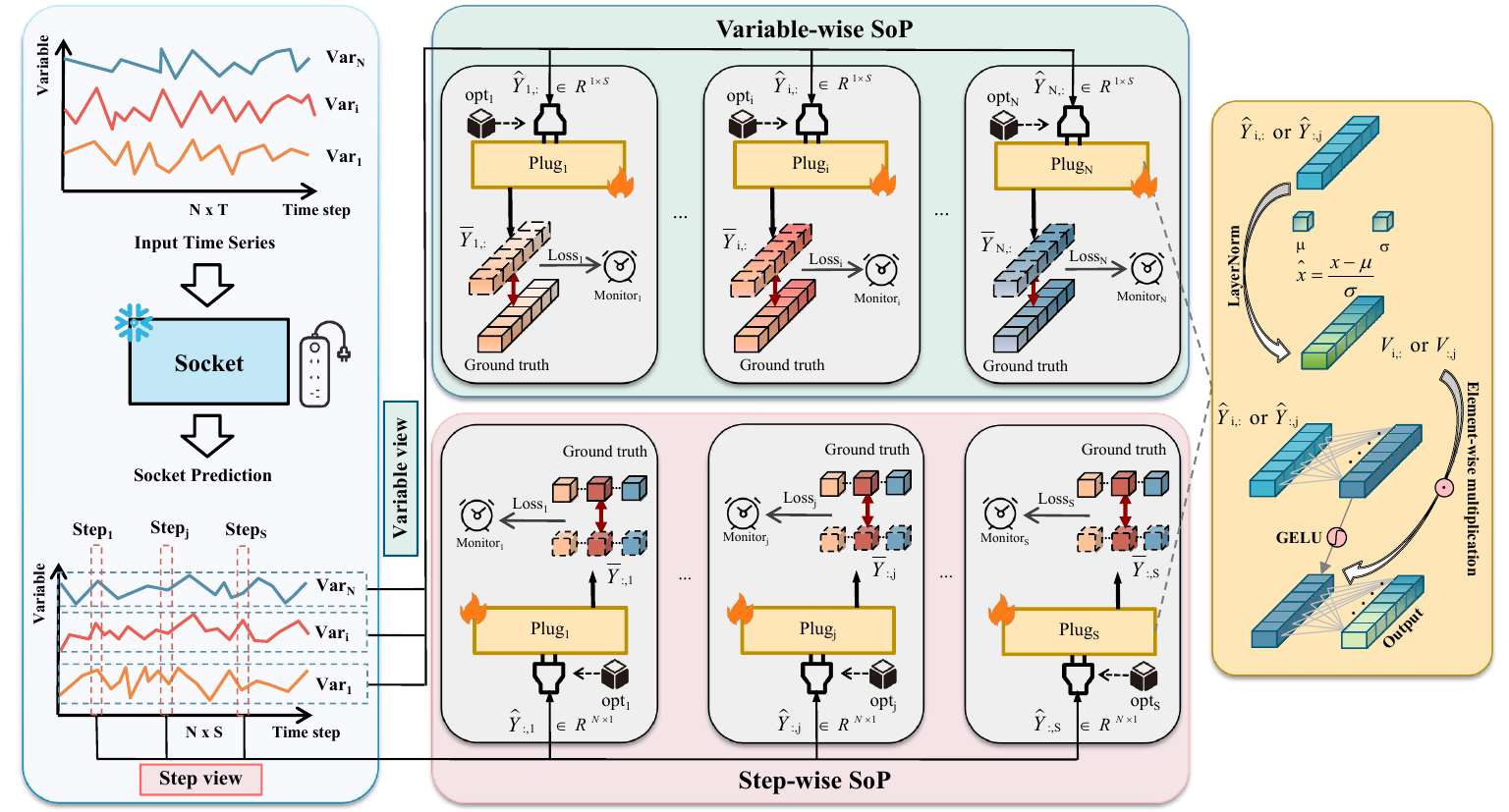}
\caption{The framework of the proposed SoP. On the left are Socket, such as fully trained models like iTransformer, DLinear, etc., which provide the foundational forecasts. On the right are independently deployed variable-wise Plug (upper part) and step-wise Plug (lower part), which aims to calibrate the forecasts produced by the Socket. 
}
\label{fig：kuangjia}
\end{figure*}

\subsection{Model Inference}

As shown in~\autoref{fig：kuangjia}, the proposed SoP strategy operates on the basis of two key components: the Socket and the Plug. The Socket is adopted to capture complex correlations among variables and deliver the high-quality foundational forecasts. To this end, it is recommended to harness {a trained SOTA forecasting model, such as a Transformer-based variant}, as the Socket. Given an input sample {$X\in\mathbb{R}^{N\times T}$}, the Socket can preliminarily infer the prediction of the future sequence $\hat{Y}$ as follows:
\begin{equation}
{\hat{Y} =  \textrm{Socket}\left ( X \right ).
} \label{eq:1} 
\end{equation}
The obtained ${\hat{Y}}$ is then divided into $M$ groups along the dimension of variables or horizons. Specifically, in the case of variable-wise SoP, ${\hat{Y}}$ is divided into $N$ groups based on each variable index $i$, forming ${\hat{Y}_{i,:}} \in \mathbb{R}^{S}$. Each ${\hat{Y}_{i,:}}$ undergoes layer normalization to produce the normalized values ${V_{i,:}}$, a technique demonstrated to improve the stability of the training convergence~\cite{lei2016layer}, and then is processed by a Plug$_i$ for calibrating.

\begin{align}
V_{i,:} &= \text{LayerNorm}\left(\hat{Y}_{i,:}\right)  \label{eq:2} \\
\bar{Y}_{i,:} &= \text{Plug}_i\left(\hat{Y}_{i,:}\right) \label{eq:3} \\
&= \text{MLP}_i\left(\hat{Y}_{i,:}\right) \odot V_{i,:}. \label{eq:4}
\end{align}

\noindent {where MLP$_i$: $\mathbb{R}^{S}$$\to$$\mathbb{R}^{d}$$\to$$\mathbb{R}^{S}$ is a projection that projects the sequence representation from $S$ to the hidden dimension $d$ and finally returns to the $S$ dimension, which consists of two $d$-dimension hidden layers and a GELU activation~\cite{hendrycks2016gaussian}. The symbol $\odot$ represents the element-wise multiplication ~\cite{han2024softs}.} Finally, the predicted outputs of the $N$ Plugs are concatenated to obtain the final forecasting results $\bar{Y}$ as follows:

\begin{equation}
{\bar{Y}= \left \{ \bar{Y}_{i,:} \mid  i\in N \right \}.
}\label{eq:5} 
\end{equation}

\subsection{Non-collective Model Calibrating}
The calibrating process through variable-wise SoP is presented in Algorithm \ref{algorithm}. 
{Let a batch of data  $\textbf{X}\in\mathbb{R}^{B\times N\times T}$ and the corresponding ground truth $\textbf{Y}\in\mathbb{R}^{B\times N\times S}$, where $B$ represents the batch size, be provided.} {The Mean Squared Error (MSE) serves as the loss function $\mathcal{L}$.} The role of each Plug$_i$ is to calibrate the preliminary forecasting results $\hat{Y}_{i,:}$ for each target variable $i$. Specifically, each Plug$_i$ is equipped with a dedicated optimizer $opt_i$ for minimizing its training loss $\mathcal{L}^\text{train}_i$ and a monitor $mnt_i$ for early-stopping associated with the validation loss $\mathcal{L}^\text{val}_i$, enabling exclusive calibrating. {An analysis of efficient parallel training is also provided and has been included in Appendix~\ref{sub:analysis}, owing to space constraints.}
\begin{algorithm}[!bh]
  \caption{Training algorithm illustrated through variable-wise SoP}
  \label{algorithm}
\setlength{\baselineskip}{4pt} 

  \begin{flushleft}
    \textbf{Input:} Trained Socket; training dataset $D_{\text{train}}$; validation dataset $D_{\text{val}}$.
  \end{flushleft}
  \begin{algorithmic}[1]
    \For{$i = 1$ to $N$}
    \State initialize the parameters of Plug$_{i}$ $\gets \theta_{i} $
    \Repeat
        \State {fetch} a batch of data {$(\textbf{X}, \textbf{Y})$ from $D_{\text{train}}$}

        \State $\hat{\textbf{Y}}=\text{Socket}(\textbf{X})$
        \State $\bar{\textbf{Y}} = \text{Plug}_{i}(\hat{\textbf{Y}})$
        \State $\mathcal{L}^\text{train}_i \leftarrow \text{MSE}(\bar{\textbf{Y}}, \textbf{Y})$
        \State update $\theta_{i}$ using optimizer $opt{_i}$ by minimizing  $\mathcal{L}^\text{train}_i$
        \If{up to the end of an epoch}
        \State snapshot Plug$_{i}$ if  $\mathcal{L}^\text{val}_i$ on $D_\text{val}$ is minimal
        \EndIf
    \Until{early-stopping of $mnt_i$ is triggered}
\EndFor
\State \textbf{Output:} $N$ snapshots of trained Plugs
\end{algorithmic}
\end{algorithm}

\section{Experiments}
In this section, we present the experimental setup and results of SoP, focusing on the following research questions:
\begin{itemize}
\item \textbf{RQ1}: How effective is SoP in enhancing the performance of existing time series forecasting models such as DLinear, TimesNet, and others? 
\item \textbf{RQ2}: How effective is the target-wise SoP, specifically in terms of variable-wise and step-wise calibration?
\item \textbf{RQ3}: {How does the performance differ between non-collective and collective optimizer deployment? }
\item \textbf{RQ4}: How effectively can SoP extend its functionality to spatio-temporal forecasting?
\end{itemize}

\subsection{Experimental Setup}\label{AA}


 
\noindent \textbf{Datasets.}
We initially conducted extensive experiments on benchmark time series datasets, including ETTh1, ETTh2, ECL, Exchange, Weather, and Solar-Energy.

\noindent \textbf{Inference models.} {SoP is a model-agnostic method that can be broadly applied to any trained deep neural networks.} We assess the effectiveness of SoP by applying it to seven SOTA forecasting methods as the inference models, also referred to as Sockets, which are categorized as follows: (1) Transformer-based methods including FEDformer~\cite{zhou2022fedformer}, PatchTST~\cite{pachtst}, and iTransformer~\cite{liu2023itransformer}; (2) MLP-based methods including DLinear~\cite{Zeng_Chen_Zhang_Xu_2023_dlnear}, TSMixer~\cite{chen2023tsmixer}, and SOFTS~\cite{han2024softs}; and (3) a TCN-based method: TimesNet~\cite{wu2022timesnet}.

\begin{table*}[htb!]
\centering
\resizebox{\linewidth}{!}{
\begin{tabular}{ccccccccccccccc}
\hline
Models &
  \multicolumn{2}{c}{ECL} &
  \multicolumn{2}{c}{Weather} &
  \multicolumn{2}{c}{Exchange} &
  \multicolumn{2}{c}{Traffic} &
  \multicolumn{2}{c}{Solar-Energy} &
  \multicolumn{2}{c}{ETTh1} &
  \multicolumn{2}{c}{ETTh2} \\ \cline{2-15} 
Metric &
  MSE &
  MAE &
  MSE &
  MAE &
  MSE &
  MAE &
  MSE &
  MAE &
  MSE &
  MAE &
  MSE &
  MAE &
  MSE &
  MAE \\ \hline
  SOFTS             & 0.175   & 0.265   & 0.259   & 0.280   & 0.413    & 0.427    & 0.406   & 0.304    & 0.234   & 0.261    & 0.456   & 0.448    & 0.383   & 0.405   \\
 SOFTS+Plug        & 0.170   & 0.263   & 0.249   & 0.273   & 0.365    & 0.430    & 0.401   & 0.265    & 0.232   & 0.258    & 0.452   & 0.447    & 0.381   & 0.403   \\
\textbf{Promotion} & \textbf{2.768\%} & \textbf{0.573\%} & \textbf{3.729\%} & \textbf{2.495\%} & \textbf{11.531\%} & -0.820\% & \textbf{1.292\%} & \textbf{12.897\%} & \textbf{1.248\%} & \textbf{1.136\%} & \textbf{1.012\%} & \textbf{0.254\%} & \textbf{0.521\%} & \textbf{0.662\%} \\
\hline
iTransformer      & 0.175   & 0.266   & 0.261   & 0.281   & 0.382    & 0.418    & 0.421   & 0.282    & 0.234   & 0.261    & 0.459   & 0.450    & 0.384   & 0.407   \\
iTransformer+Plug & 0.169   & 0.264   & 0.249   & 0.274   & 0.295    & 0.401    & 0.415   & 0.279    & 0.231   & 0.260    & 0.450   & 0.447    & 0.381   & 0.404   \\
\textbf{Promotion} & \textbf{3.413\%} & \textbf{0.902\%} & \textbf{4.795\%} & \textbf{2.598\%} & \textbf{22.907\%} & \textbf{4.182\%} & \textbf{1.431\%} & \textbf{1.112\%} & \textbf{1.370\%} & \textbf{0.384\%} & \textbf{1.892\%} & \textbf{0.713\%} & \textbf{0.671\%} & \textbf{0.652\%} \\
\hline
TimesNet          & 0.195   & 0.296   & 0.261   & 0.287   & 0.422    & 0.445    & 0.629   & 0.333    & 0.263   & 0.274    & 0.477   & 0.466    & 0.413   & 0.426   \\
TimesNet+Plug     & 0.190   & 0.290   & 0.261   & 0.286   & 0.338    & 0.422    & 0.598   & 0.332    & 0.262   & 0.274    & 0.473   & 0.465    & 0.410   & 0.420   \\
\textbf{Promotion} & \textbf{2.684\%} & \textbf{1.808\%} & \textbf{0.019\%} & \textbf{0.415\%} & \textbf{19.986\%} & \textbf{5.239\%} & \textbf{4.981\%} & \textbf{0.207\%} & \textbf{0.560\%} & -0.301\% & \textbf{0.814\%} & \textbf{0.139\%} & \textbf{0.785\%} & \textbf{1.371\%} \\
\hline
PatchTST          & 0.204   & 0.294   & 0.256   & 0.279   & 0.390    & 0.417    & 0.464   & 0.296    & 0.280   & 0.313    & 0.452   & 0.450    & 0.398   & 0.417   \\
PatchTST+Plug     & 0.184   & 0.277   & 0.248   & 0.274   & 0.306    & 0.389    & 0.454   & 0.292    & 0.271   & 0.300    & 0.450   & 0.453    & 0.380   & 0.408   \\
\textbf{Promotion} & \textbf{9.852\%} & \textbf{5.792\%} & \textbf{3.303\%} & \textbf{1.837\%} & \textbf{21.639\%} & \textbf{6.781\%} & \textbf{2.073\%} & \textbf{1.084\%} & \textbf{3.059\%} & \textbf{3.941\%} & \textbf{0.282\%} & -0.735\% & \textbf{4.590\%} & \textbf{2.306\%} \\
\hline
FEDformer        & 0.229   & 0.340   & 0.311   & 0.359   & 0.518    & 0.508    & 0.611   & 0.378    & 0.316   & 0.393    & 0.443   & 0.457    & 0.433   & 0.449   \\
FEDformer+Plug    & 0.207   & 0.320   & 0.299   & 0.348   & 0.404    & 0.470    & 0.603   & 0.370    & 0.289   & 0.367    & 0.433   & 0.451    & 0.425   & 0.441   \\
\textbf{Promotion} & \textbf{9.733\%} & \textbf{5.689\%} & \textbf{3.822\%} & \textbf{3.272\%} & \textbf{22.040\%} & \textbf{7.493\%} & \textbf{1.307\%} & \textbf{2.297\%} & \textbf{8.571\%} & \textbf{6.665\%} & \textbf{2.339\%} & \textbf{1.441\%} & \textbf{1.838\%} & \textbf{1.915\%} \\
\hline

DLinear            & 0.213   & 0.302   & 0.265   & 0.316   & 0.341    & 0.402    & 0.672   & 0.419    & 0.330   & 0.401    & 0.461   & 0.457    & 0.565   & 0.521   \\
DLinear+Plug        & 0.182   & 0.281   & 0.241   & 0.295   & 0.290    & 0.387    & 0.577   & 0.361    & 0.256   & 0.310    & 0.451   & 0.453    & 0.548   & 0.508   \\
\textbf{Promotion} & \textbf{14.396\%} & \textbf{6.962\%} & \textbf{8.966\%} & \textbf{6.779\%} & \textbf{13.682\%} & \textbf{3.574\%} & \textbf{14.133\%} & \textbf{13.775\%} & \textbf{22.364\%} & \textbf{22.719\%} & \textbf{2.146\%} & \textbf{0.852\%} & \textbf{3.058\%} & \textbf{2.387\%} \\
\hline
\end{tabular}
}
\caption{The averaged performance across all prediction horizons \({S = \{96, 192, 336, 720\}}\) comparing the inference model and its enhancement with variable-wise SoP.}
\label{tab:my-table1}
\end{table*}

\noindent \textbf{Experimental settings.}
 We utilized the open library TSLib~\cite{tslib}, which enabled the easy reproduction of aforementioned inference models. For SOFTS, which is not included in TSLib, we reproduced it using the hyperparameter settings provided in the original paper~\cite{han2024softs}. {In the Appendix, we provide details of the time series data set (Section~\ref{Time series dataset}), inference models (Section~\ref{baseline_models}), and the hyperparameter settings (Section~\ref{timeHyperparameter settings}).}


\subsection{Effectiveness of SoP to Enhance Inference Models (\textbf{address RQ1})}
Table~\ref{tab:my-table1} reports the averaged performance across all horizons, comparing the inference models and their enhancement using variable-wise SoP across seven benchmark datasets. The results demonstrate that SoP significantly enhances forecasting accuracy across various baselines, whether MLP- or Transformer-based methods. Notably, all Model+Plug methods achieve an improvement rate exceeding 10\% in MSE on the Exchange dataset. 
Based on prior findings~\cite{li2024foundts}, which identified datasets like Exchange as having lower inter-variable correlations and datasets like Weather as having higher inter-variable correlations, we conclude that {variable-wise} SoP performs better with datasets characterized by lower inter-variable correlations than with those exhibiting higher inter-variable correlations. This also implies the importance of understanding dataset characteristics to optimize the potential of SoP further. 
{Notably, while DLinear underperforms compared to TimeNet, the integration of SoP enables DLinear+Plug surpasses TimesNet+Plug on most datasets.}  


\begin{figure}[t]
  \centering
  \includegraphics[width=0.95\linewidth]{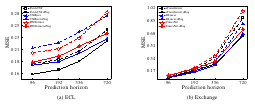}  \caption{{Performance of SoP at different prediction horizons.}}
  \label{fig：ecl-exchange}
\end{figure}

~\autoref{fig：ecl-exchange} illustrates the performance of applying SoP to each of the three models with prediction horizons ${S = \left \{ 96, 192, 336, 720 \right \}}$ on the ECL and Exchange datasets. The experimental results demonstrate that the SoP strategy consistently decreases the MSE at diverse prediction horizons. As the prediction horizon expands, the performance improvement brought by SoP becomes increasingly evident, manifested by the growing difference between the solid and dashed lines of the same color. {Experiments on Weather and Traffic datasets also demonstrates the superior performance of SoP (Section Appendix.~\ref{different prediction horizons}) and achieves competitive performance compared with LLM methods for time series forecasting (Section Appendix.~\ref{sub:llm}).}

\subsection{Effectiveness of Target-wise SoP (\textbf{address RQ2})}
To investigate the effectiveness of target-wise SoP, we examine the performance SoP with different plug counts on the Weather dataset, which contains a total of 21 target variables. {This experiment follows equitable settings for parameter size of plugs to ensure a fair fitting capability (details in Appendix.~\ref{sub:count})}.

\begin{figure}[!ht]
  \centering
  \includegraphics[width=\linewidth]{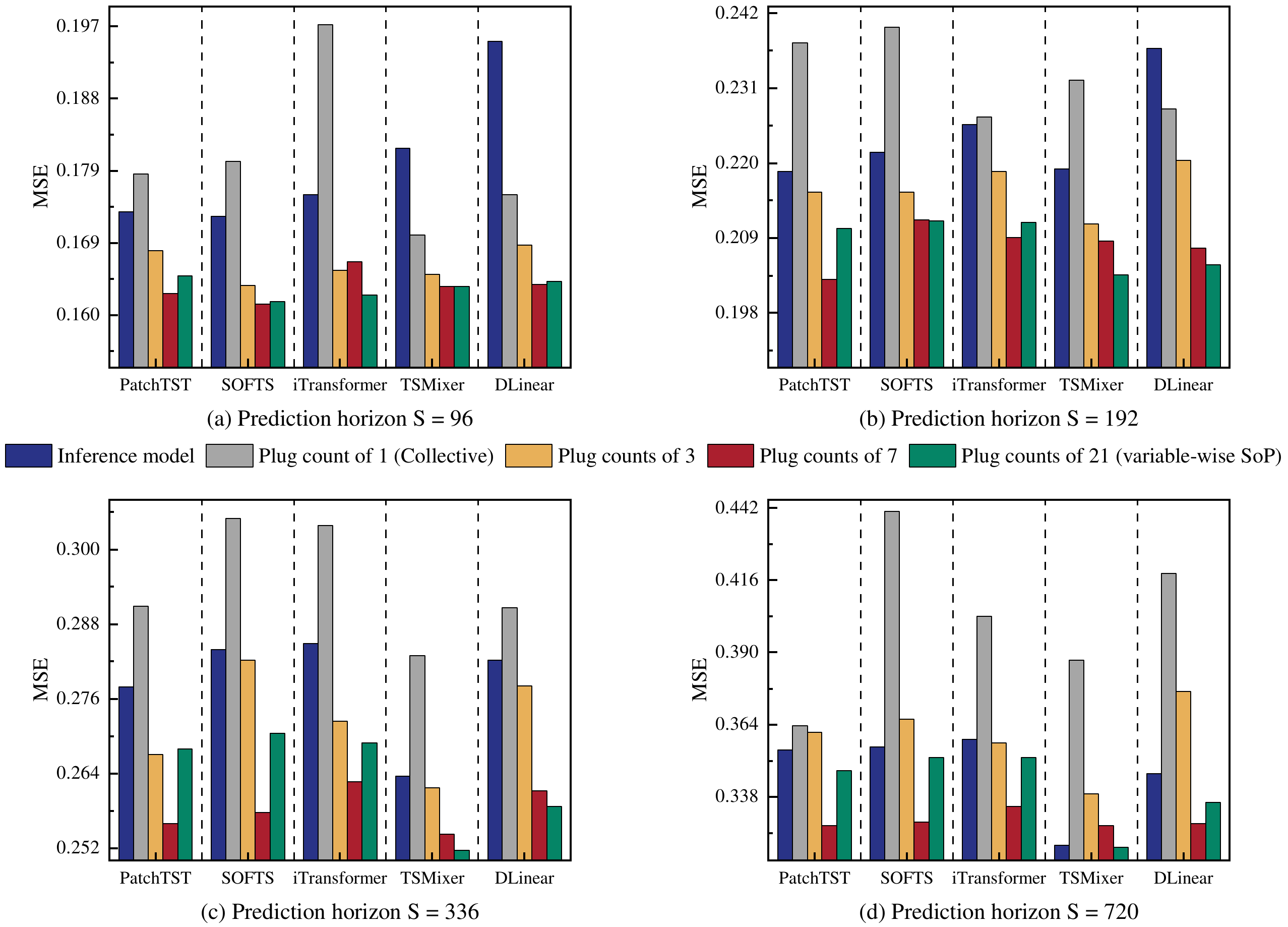}
  \caption{Effect of plug counts on the performance of SoP from the variable view.}
  \label{fig:exchange}
\end{figure}

{\textit{Effect of SoP from Variable View}. Given a total of 21 target variables, we specifically setup the plug counts along the $N$-dimension to form \(\{21, 7, 3, 1\}\) plugs in turn}. The experimental results, shown in \autoref{fig:exchange}, reveal several key observations. First, most non-collective SoP clearly enhances the performance of inference models, with SoP using plug count of 7 achieving the best performance, followed closely by variable-wise SoP. The variable-wise SoP, by eliminating the need to fine-tune the plug count, is thus suggested as a quick-start version of SoP. Second, plug count of 1 (i.e., collective calibrating) generally results in the worst performance, even performing worse than the inference model without calibrating. This suggests the MTLC indeed degrades the performance of most models. Lastly, models with simpler architectures, such as DLinear, benefit more from incorporating SoP.

\begin{table*}[!ht]
\centering
\resizebox{\linewidth}{!}{
\begin{tabular}{cccccccccccccccc}
\hline
\multicolumn{2}{c}{Model of Socket} &
  \multicolumn{2}{c}{SOFTS} &
  \multicolumn{2}{c}{iTransformer} &
  \multicolumn{2}{c}{DLinear} &
  \multicolumn{2}{c}{PatchTST
  } &
  \multicolumn{2}{c}{TimesNet} &
  \multicolumn{2}{c}{TSMixer} &
  \multicolumn{2}{c}{FEDformer} \\ \cline{3-16} 
\multicolumn{2}{c}{Metric} &
  MSE &
  MAE &
  MSE &
  MAE &
  MSE &
  MAE &
  MSE &
  MAE &
  MSE &
  MAE &
  MSE &
  MAE &
  MSE &
  MAE \\ \hline
 &
  {\color[HTML]{333333} Promotion1} &
   \textbf{2.77\%} &
   \textbf{0.57\%} &
   \textbf{3.41\%} &
   \textbf{0.90\%} &
   \textbf{14.40\%} &
   \textbf{6.96\%} &
   \textbf{9.85\%} &
   \textbf{5.79\%} &
  2.68\% &
   \textbf{1.81\%} &
   \textbf{16.29\%} &
   \textbf{9.80\%} &
   \textbf{9.73\%} &
   \textbf{5.69\%} \\
\multirow{-2}{*}{ECL} &
  {\color[HTML]{333333} Promotion2} &
  1.92\% &
  0.05\% &
  3.28\% &
  0.40\% &
  12.72\% &
  6.20\% &
  9.27\% &
  5.54\% &
  \textbf{3.23\%} &
  1.22\% &
  15.59\% &
  9.40\% &
  9.52\% &
  5.55\% \\ \hline
 &
  {\color[HTML]{333333} Promotion1} &
  \textbf{3.73\%} &
  \textbf{2.49\%} &
  \textbf{4.80\%} &
  \textbf{2.60\%} &
  \textbf{8.97\%} &
  \textbf{6.78\%} &
  \textbf{3.30\%} &
  \textbf{1.84\%} &
  \textbf{0.02\%} &
  \textbf{0.42\%} &
  \textbf{4.69\%} &
  \textbf{3.47\%} &
  3.82\% &
  3.27\% \\
\multirow{-2}{*}{Weather} &
  {\color[HTML]{333333} Promotion2} &
  2.72\% &
  1.41\% &
  4.20\% &
  2.04\% &
  5.84\% &
  5.11\% &
  2.49\% &
  1.23\% &
  -0.14\% &
  -0.06\% &
  3.59\% &
  2.41\% &
  \textbf{4.56\%} &
  \textbf{5.69\%} \\ \hline
 &
  {\color[HTML]{333333} Promotion1} &
  \textbf{11.53\%} &
  \textbf{-0.82\%} &
  \textbf{22.91\%} &
  \textbf{4.18\%} &
  \textbf{14.68\%} &
  \textbf{3.57\%} &
  \textbf{21.64\%} &
  \textbf{6.78\%} &
  \textbf{19.99\%} &
  \textbf{5.24\%} &
  \textbf{1.37\%} &
  \textbf{1.48\%} &
  \textbf{22.04\%} &
  \textbf{7.49\%} \\
\multirow{-2}{*}{Exchange} &
  {\color[HTML]{333333} Promotion2} &
  7.03\% &
  -0.89\% &
  21.99\% &
  6.10\% &
  4.49\% &
  -3.40\% &
  18.60\% &
  3.06\% &
  15.72\% &
  2.86\% &
  -5.10\% &
  -1.56\% &
  21.22\% &
  7.39\% \\ \hline
 &
  {\color[HTML]{333333} Promotion1} &
  \textbf{1.29\%} &\textbf{12.90\%} &
\textbf{1.43\%} &
  \textbf{1.11\%} &
  \textbf{14.13\%} &
  \textbf{13.78\%} &
  \textbf{2.07\%} &
  \textbf{1.08\%} &
  \textbf{4.98\%} &
  \textbf{0.21\%} &
  1.79\% &
  3.24\% &
  \textbf{1.31\%} &
  \textbf{2.30\%} \\
\multirow{-2}{*}{Traffic} &
  {\color[HTML]{333333} Promotion2} &
  0.56\% &
  12.85\% &
  0.91\% &
  0.84\% &
  12.33\% &
  15.48\% &
  0.71\% &
  -1.27\% &
  0.73\% &
  -0.40\% &
  \textbf{ 1.85\%} &
  \textbf{ 3.91\%} &
  1.14\% &
  2.09\% \\ \hline
 &
  {\color[HTML]{333333} Promotion1} &
  \textbf{1.25\%} &
  \textbf{1.14\%} &
  \textbf{1.37\%} &
  \textbf{0.38\%} &
  \textbf{22.36\%} &
  \textbf{22.72\%} &
  \textbf{3.06\%} &
  \textbf{3.97\%} &
  \textbf{0.56\%} &
  \textbf{-0.30\%} &
  \textbf{0.64\%} &
  \textbf{0.02\%} &
  \textbf{8.57\%} &
  \textbf{6.66\%} \\
\multirow{-2}{*}{Solar-Energy} &
  {\color[HTML]{333333} Promotion2} &
  1.12\% &
  0.31\% &
  1.22\% &
  0.23\% &
  21.62\% &
  22.47\% &
  2.87\% &
  2.79\% &
  -0.53\% &
  -6.06\% &
  0.63\% &
  -0.62\% &
  8.10\% &
  6.47\% \\ \hline
\multicolumn{2}{c}{{\color[HTML]{333333} Avg Promotion1}} &
  {\color[HTML]{FE0000} \textbf{3.16\%}} &
  {\color[HTML]{FE0000} \textbf{2.46\%}} &
  {\color[HTML]{FE0000} \textbf{5.21\%}} &
  {\color[HTML]{FE0000} \textbf{4.77\%}} &
  {\color[HTML]{FE0000} \textbf{14.91\%}} &
  {\color[HTML]{FE0000} \textbf{11.40\%}} &
  {\color[HTML]{FE0000} \textbf{6.40\%}} &
  {\color[HTML]{FE0000} \textbf{4.82\%}} &
  {\color[HTML]{FE0000} \textbf{4.26\%}} &
  {\color[HTML]{FE0000} \textbf{2.58\%}} &
  {\color[HTML]{FE0000} \textbf{4.95\%}} &
  {\color[HTML]{FE0000} \textbf{3.32\%}} &
  {\color[HTML]{FE0000} \textbf{7.09\%}} &
  {\color[HTML]{FE0000} \textbf{6.88\%}} \\
\multicolumn{2}{c}{{\color[HTML]{333333} Avg Promotion2}} &
  2.11\% &
  2.05\% &
  1.51\% &
  1.59\% &
  10.76\% &
  9.17\% &
  3.00\% &
  1.31\% &
  1.27\% &
  -0.24\% &
  3.60\% &
  2.71\% &
  4.11\% &
  4.33\% \\ \hline
\end{tabular}
}
\caption{The accuracy improvements achieved by non-collective SoP (Promotion1) and collective calibrating (Promotion2).}
\label{tab:my-table5}
\end{table*}

\begin{figure}[!ht]
\includegraphics[width=0.93\linewidth]{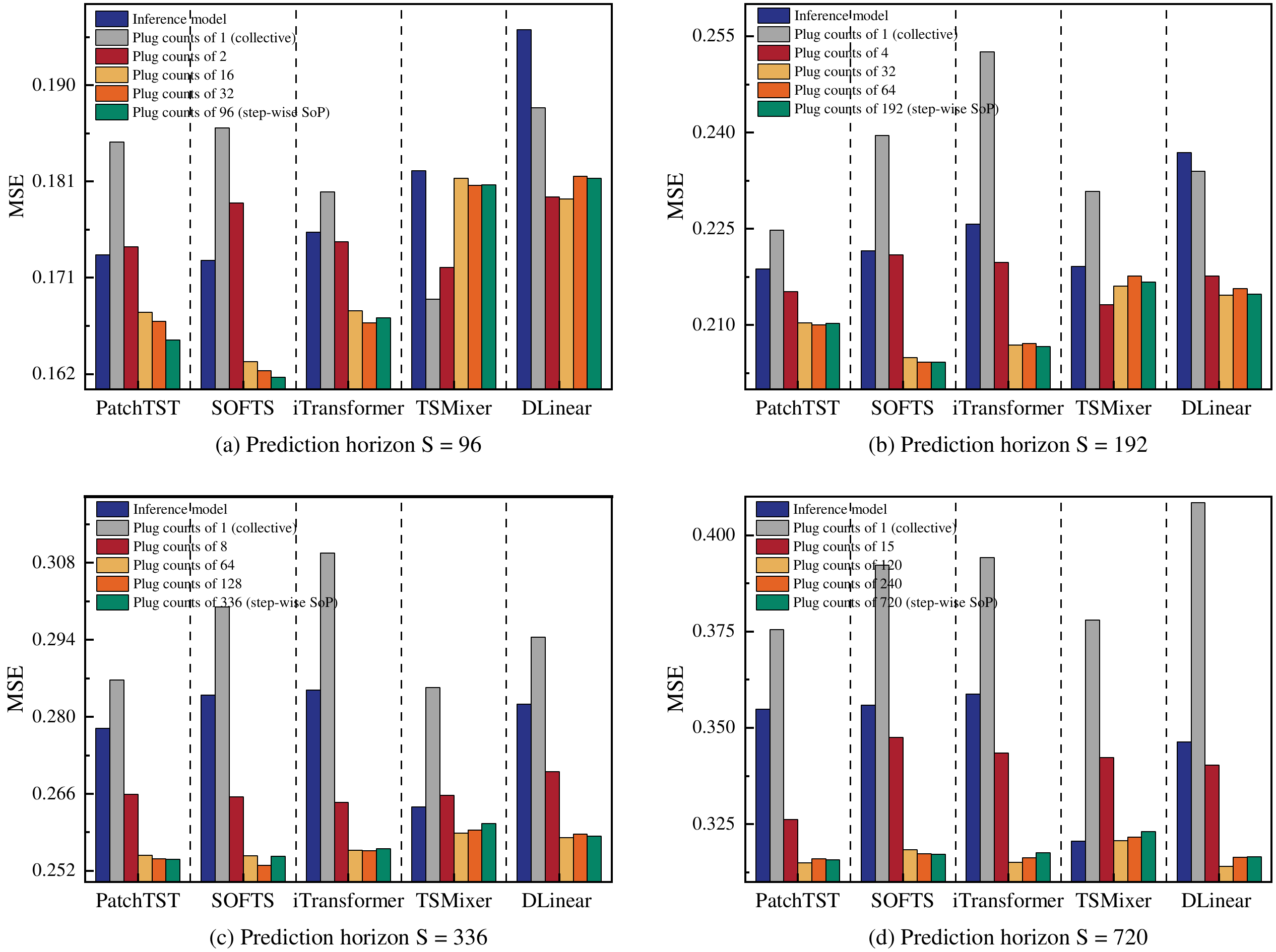}
    \caption{Effect of plug counts on the performance of SoP from the step view.}
    \label{fig：weather}
\end{figure}

\textit{Effect of SoP from Step View.} Given that \( S \in \{96, 192, 336, 720\} \), we set up the plug counts along the \( S \)-dimension as follows: for \( S=96\), the plug counts form \(\{96, 32, 16, 2, 1\}\); for \(S=192\) , the plug counts form \(\{192, 64, 32, 4, 1\}\); for \( S=336 \), the plug counts form \(\{336, 128, 64, 8, 1\}\); and for \( S=720 \), the plug counts form \(\{720, 240, 120, 15, 1\}\). As shown in \autoref{fig：weather}, the experimental results reveal that MTLC degrades the performance of most models, except for TSMixer, which performs best with collective calibration when \( S=96 \). In contrast, non-collective SoP significantly enhances model performance, with PatchTST, implementing step-wise SoP, achieving the best results across most prediction horizons. Overall, the step-wise SoP, by eliminating the need to experiment with specific plug counts, performs considerably well and can be chosen as a quick-start version of SoP from the step view.

\subsection{Ablation Study of Non-collective and Collective Optimizer Deployment (\textbf{address RQ3})}
As highlighted by the MTLC phenomenon in~\autoref{fig：haibao}, we conducted a more rigorous ablation study to compare the effects of non-collective and collective optimizer deployment. {Experimental results demonstrate the number of training epochs required to trigger early-stopping of two deployments are significantly distinct (Section Appendix.~\ref{sub:collective}.)} 

Table \ref{tab:my-table5} summarizes the performance achieved by deploying two different calibrating strategies. The improvements are calculated relative to the performance of the Socket without calibrating, with the superior improvement in each comparison highlighted in bold. Promotion 1 is achieved through variable-wise SoP, where each plug is optimized independently by a specific optimizer. Promotion 2 is achieved via collective calibrating, where all plugs share a unified optimizer. The results show that while both calibrating approaches yield performance improvements, SoP leads to more substantial gains. The average improvement achieved by Promotion 1 is approximately twice that of Promotion 2. {A striking example of this difference can be observed with the TSMixer model on the Exchange dataset, where collective calibrating results in a degradation in model performance, while the corresponding SoP strategy leads to significant improvements.} Our experiments also exhibited the SoP with considerable robustness (Section Appendix.~\ref{sub:robust}) and transferability of the trained plug (Section Appendix.~\ref{sub:transfer}).

\subsection{Experiments on Spatio-temporal Forecasting (\textbf{address RQ4})}

Both time series forecasting and spatio-temporal forecasting aim to predict future values based on historical data, but spatio-temporal forecasting is more complex in terms of data scale and model complexity. This section demonstrates the effectiveness of SoP in spatio-temporal forecasting.


\noindent \textbf{Dataset.} The experiments were conducted on the ERA5 dataset~\cite{tianchi21}, which includes five meteorological variables: T2M, U10, V10, MSL and TP. 


\noindent \textbf{Inference model.} The inference model selected as the Socket is Unet~\cite{ronneberger2015u}. A direct prediction strategy is employed, which generates predictions for the next 20 prediction horizons without iterative steps. 

\noindent \textbf{Experimental settings.} 
For spatio-temporal prediction, a data sample is represented as \(X \in \mathbb{R}^{N \times H \times W \times T}\), with \(Y \in \mathbb{R}^{N \times H \times W \times S}\) as the corresponding output. In this forecasting task, observations from the past two time steps (\(T = 2\)) involving five variables (\(N = 5\)) are used to predict the same five variables over the next 20 prediction horizons (\(S = 20\)). \(H \times W\) denotes the spatial range, with \(W = H = 160\). {The details on dataset partitioning and the parameter settings for Unet are provided in Appendix~\ref{Spatio-temporal dataset}.}




\noindent \textbf{Experimental results.} Next, we provide the analysis of the experimental results in detail.

\begin{figure}[htb!]
  \centering
  \includegraphics[width=0.9\linewidth]{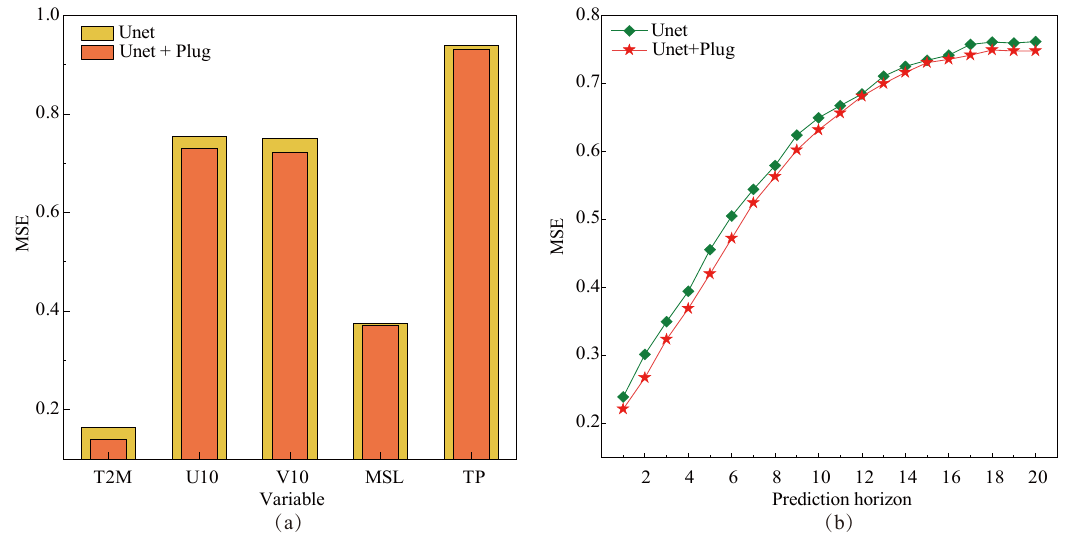}
  \caption{(a) The averaged MSE for each of five variables. (b) The averaged MSE at each prediction horizon.}
  \label{fig：55555}
\end{figure}

\begin{figure}[htb!]
  \centering
  \includegraphics[width=\linewidth]{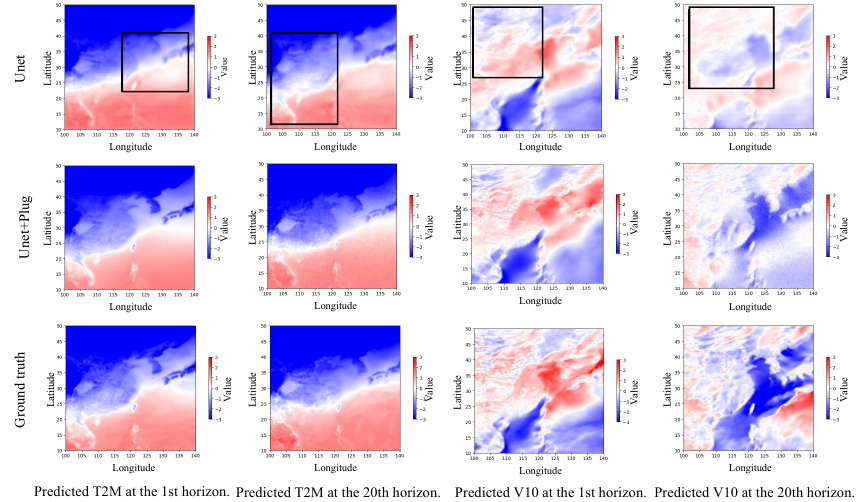}
  \caption{Visualization of a spatio-temporal forecasting case study for prediction horizons $S$=1 and $S$=20, as predicted by Unet (top), Unet+Plug (middle), and the ground truth (bottom). Columns 1–2 and 3–4 display T2M and V10, respectively.}
  \label{fig：SK}
\end{figure}

\textit{The Effect of SoP on Spatio-temporal Variables.} As shown in~\autoref{fig：55555} (a), the averaged test MSE of the five variables significantly decreased after applying SoP to Unet, particularly for the variables T2M, U10, and V10. This improvement can be attributed to the inherently stronger cyclicity of T2M, which enhances its predictability compared to TP and MSL, which aligns with findings in previous studies~\cite{liu2021evaluation}.~\autoref{fig：SK} displays a test sample to visualize the benefits of applying SoP to the Unet model for spatio-temporal predictions of T2M and V10. It can be observed that predictions from Unet alone appear overly smoothed and obscurer at certain spatial locations (highlighted in the black box). In contrast, Unet+SoP produces more textured and sharper predictions that are closer to the ground truth, not only at the first prediction horizon but also at the 20th prediction horizon.  

\textit{The Effect of SoP on  Prediction Horizons.} As present in~\autoref{fig：55555} (b), the incorporation of the SoP consistently reduces the test MSE across each of the 20 prediction horizons. Notably, the difference in MSE between Unet+Plug and Unet gradually decreases as the prediction horizon increase. This phenomenon can be attributed to the reduced predictability of spatio-temporal forecasts as the horizon progresses~\cite{bi2022pangu}. {We further display the prediction performance with a case study in Appendix.~\ref{sub:variables}.} Overall, SoP demonstrates its effectiveness both in short- and long-term spatio-temporal forecasting.  

\section{Related Works}
{\paragraph{Forecasting Models.} Deep learning has significantly advanced time series forecasting, resulting in the development of numerous deep forecasting models ~\cite{deng2024parsimony}. 
The architecture of deep models has progressed from CNNs-based frameworks~\cite{wu2022timesnet}, MLPs~\cite{Zeng_Chen_Zhang_Xu_2023_dlnear} to Transformer~\cite{liu2023itransformer}, achieving considerable performance improvements across various benchmarks. Recently, the rise of LLMs has opened new possibilities for time series forecasting~\cite{liu2024unitime,zhong2025time}. However, some studies have shown that these popular LLMs often perform similarly to, or even worse than, simpler methods, while also significantly increasing computational costs during both training and inference~\cite{tan2024language}. As a result, there is no universally superior model architecture in the research community of time series forecasting. 


\paragraph{Forecasting Strategies.}
Recent studies have shown that well-designed forecasting strategies—such as multi-view learning~\cite{deng2022multi,lu2019fuzzy} and model calibration—can significantly influence predictive performance. Calibration, in particular, plays a critical role in aligning model predictions with the ground truth~\cite{ZhangMDZ23}, or in providing more reliable confidence intervals~\cite{SunLL23}. Early research employed Platt Scaling to transform the original predictions into calibrated probabilities~\cite{platt1999probabilistic}. More recently, non-parametric isotonic regression was introduced by~\cite{berta2024classifier} to calibrate binary classifiers.~\cite{Correction23} designed an abductive reasoning mechanism to minimize the discrepancy between the inferred and the true values to adjust the predictions.~\cite{ZhangMDZ23} employed the envelope-based bounds modeling to correct the initial predictions. While many previous studies have applied learning-based methods to calibrate model outputs, most have focused on collective calibrating. None of these approaches have noticed the detriment of MTLC in deep forecasting and proposed a solution through non-collective calibrating. To this end, this study has developed an effective calibrating strategy from two views that enhances the performance of deep forecasting models.

\section{Conclusions}
This study identifies a detrimental phenomenon, termed MTLC, which hampers the learning capability during the training of deep forecasting models. To address this issue, we propose an effective SoP strategy that calibrates well-trained forecasting models, regardless of their specific architectures. Additionally, we derive two specific forms of target-wise SoP for quick-start implementation. Comprehensive experiments demonstrate the superiority of the proposed method, even without exhaustive hyperparameter searching. We hope this approach paves the way for further advancements in time series forecasting. Future research could focus on improving calibrating efficiency through parallel Plug training.

\section{Acknowledgments}
 This work was supported by the National Natural Science Foundation of China (Nos. 62402463, 41976185, 62176243, 62176221 and 72242106) and Major Basic Research Project of Shandong Provincial Natural Science Foundation (ZR2024ZD03). 

\bibliographystyle{named}
\bibliography{main}

\begin{thebibliography}{}

\bibitem[\protect\citeauthoryear{{Alibaba Group}}{2023}]{tianchi21}
{Alibaba Group}.
\newblock {The First World Science and Intelligence Competition: Atmospheric Science Track - East China AI Medium-Range Weather Forecasting Competition}.
\newblock {\url{https://tianchi.aliyun.com/competition/entrance/532111/information}}, 2023.

\bibitem[\protect\citeauthoryear{Ba \bgroup \em et al.\egroup }{2016}]{lei2016layer}
Lei~Jimmy Ba, Jamie~Ryan Kiros, and Geoffrey~E. Hinton.
\newblock Layer normalization.
\newblock {\em CoRR}, abs/1607.06450, 2016.

\bibitem[\protect\citeauthoryear{Berta \bgroup \em et al.\egroup }{2024}]{berta2024classifier}
Eugene Berta, Francis Bach, and Michael Jordan.
\newblock Classifier calibration with roc-regularized isotonic regression.
\newblock In {\em International Conference on Artificial Intelligence and Statistics}, pages 1972--1980. PMLR, 2024.

\bibitem[\protect\citeauthoryear{Bi \bgroup \em et al.\egroup }{2023}]{bi2022pangu}
Kaifeng Bi, Lingxi Xie, Hengheng Zhang, Xin Chen, Xiaotao Gu, and Qi~Tian.
\newblock Accurate medium-range global weather forecasting with 3d neural networks.
\newblock {\em Nat.}, 619(7970):533--538, 2023.

\bibitem[\protect\citeauthoryear{Borovykh \bgroup \em et al.\egroup }{2017}]{2017Conditional}
Anastasia Borovykh, Sander Bohte, and Cornelis~W Oosterlee.
\newblock Conditional time series forecasting with convolutional neural networks.
\newblock {\em arXiv}, 2017.

\bibitem[\protect\citeauthoryear{Chen \bgroup \em et al.\egroup }{2023a}]{chen2023fuxi}
Lei Chen, Xiaohui Zhong, Feng Zhang, Yuan Cheng, Yinghui Xu, Yuan Qi, and Hao Li.
\newblock {FuXi}: A cascade machine learning forecasting system for 15-day global weather forecast.
\newblock {\em npj Climate and Atmospheric Science}, 6(1):190, 2023.

\bibitem[\protect\citeauthoryear{Chen \bgroup \em et al.\egroup }{2023b}]{chen2023tsmixer}
Si{-}An Chen, Chun{-}Liang Li, Sercan~{\"{O}}. Arik, Nathanael~C. Yoder, and Tomas Pfister.
\newblock {TSMixer}: An all-mlp architecture for time series forecast-ing.
\newblock {\em Trans. Mach. Learn. Res.}, 2023, 2023.

\bibitem[\protect\citeauthoryear{Deng \bgroup \em et al.\egroup }{2022}]{deng2022multi}
Jinliang Deng, Xiusi Chen, Renhe Jiang, Xuan Song, and Ivor~W Tsang.
\newblock A multi-view multi-task learning framework for multi-variate time series forecasting.
\newblock {\em IEEE Transactions on Knowledge and Data Engineering}, 35(8):7665--7680, 2022.

\bibitem[\protect\citeauthoryear{Deng \bgroup \em et al.\egroup }{2024a}]{deng2024disentangling}
Jinliang Deng, Xiusi Chen, Renhe Jiang, Du~Yin, Yi~Yang, Xuan Song, and Ivor~W. Tsang.
\newblock Disentangling structured components: Towards adaptive, interpretable and scalable time series forecasting.
\newblock {\em IEEE Transactions on Knowledge and Data Engineering}, 2024.

\bibitem[\protect\citeauthoryear{Deng \bgroup \em et al.\egroup }{2024b}]{deng2024parsimony}
Jinliang Deng, Feiyang Ye, Du~Yin, Xuan Song, Ivor Tsang, and Hui Xiong.
\newblock Parsimony or capability? decomposition delivers both in long-term time series forecasting.
\newblock {\em Advances in Neural Information Processing Systems}, 37:66687--66712, 2024.

\bibitem[\protect\citeauthoryear{Han \bgroup \em et al.\egroup }{2024}]{han2024softs}
Lu~Han, Xu{-}Yang Chen, Han{-}Jia Ye, and De{-}Chuan Zhan.
\newblock {SOFTS:} efficient multivariate time series forecasting with series-core fusion.
\newblock {\em CoRR}, abs/2404.14197, 2024.

\bibitem[\protect\citeauthoryear{Hendrycks and Gimpel}{2016}]{hendrycks2016gaussian}
Dan Hendrycks and Kevin Gimpel.
\newblock Gaussian error linear units (gelus).
\newblock {\em arXiv preprint arXiv:1606.08415}, 2016.

\bibitem[\protect\citeauthoryear{Huang \bgroup \em et al.\egroup }{2023}]{Correction23}
Yulong Huang, Yang Zhang, Qifan Wang, Chenxu Wang, and Fuli Feng.
\newblock Prediction then correction: An abductive prediction correction method for sequential recommendation.
\newblock In {\em {SIGIR}}, pages 2272--2276. {ACM}, 2023.

\bibitem[\protect\citeauthoryear{Jin \bgroup \em et al.\egroup }{2024}]{timellm}
Ming Jin, Shiyu Wang, Lintao Ma, Zhixuan Chu, James~Y. Zhang, Xiaoming Shi, Pin{-}Yu Chen, Yuxuan Liang, Yuan{-}Fang Li, Shirui Pan, and Qingsong Wen.
\newblock {Time-LLM}: Time series forecasting by reprogramming large language models.
\newblock In {\em {ICLR}}. OpenReview.net, 2024.

\bibitem[\protect\citeauthoryear{Kuleshov \bgroup \em et al.\egroup }{2018}]{KuleshovFE18}
Volodymyr Kuleshov, Nathan Fenner, and Stefano Ermon.
\newblock Accurate uncertainties for deep learning using calibrated regression.
\newblock In {\em {ICML}}, volume~80 of {\em Proceedings of Machine Learning Research}, pages 2801--2809. {PMLR}, 2018.

\bibitem[\protect\citeauthoryear{Lai \bgroup \em et al.\egroup }{2018}]{LSTNET18}
Guokun Lai, Wei{-}Cheng Chang, Yiming Yang, and Hanxiao Liu.
\newblock Modeling long- and short-term temporal patterns with deep neural networks.
\newblock In {\em {SIGIR}}, pages 95--104. {ACM}, 2018.

\bibitem[\protect\citeauthoryear{Liu \bgroup \em et al.\egroup }{2021}]{liu2021evaluation}
H~Liu, L~Dong, R~Yan, X~Zhang, C~Guo, S~Liang, J~Tu, X~Feng, and X~Wang.
\newblock Evaluation of near-surface wind speed climatology and long-term trend over china’s mainland region based on {ERA5} reanalysis.
\newblock {\em Climatic and Environmental Research}, 26(3):299--311, 2021.

\bibitem[\protect\citeauthoryear{Liu \bgroup \em et al.\egroup }{2024a}]{liu2024calf}
P~Liu, H~Guo, T~Dai, N~Li, J~Bao, X~Ren, Y~Jiang, and ST~Xia.
\newblock Calf: Aligning llms for time series forecasting via cross-modal fine-tuning.
\newblock {\em arXiv preprint arXiv:2403.07300}, 2024.

\bibitem[\protect\citeauthoryear{Liu \bgroup \em et al.\egroup }{2024b}]{liu2024unitime}
Xu~Liu, Junfeng Hu, Yuan Li, Shizhe Diao, Yuxuan Liang, Bryan Hooi, and Roger Zimmermann.
\newblock Unitime: A language-empowered unified model for cross-domain time series forecasting.
\newblock In {\em Proceedings of the ACM Web Conference 2024}, pages 4095--4106, 2024.

\bibitem[\protect\citeauthoryear{Liu \bgroup \em et al.\egroup }{2024c}]{liu2023itransformer}
Yong Liu, Tengge Hu, Haoran Zhang, Haixu Wu, Shiyu Wang, Lintao Ma, and Mingsheng Long.
\newblock i{T}ransformer: Inverted {T}ransformers {A}re {E}ffective for {T}ime {S}eries {F}orecasting.
\newblock In {\em {ICLR}}, 2024.

\bibitem[\protect\citeauthoryear{Lu \bgroup \em et al.\egroup }{2018}]{lu2018structural}
Jie Lu, Junyu Xuan, Guangquan Zhang, and Xiangfeng Luo.
\newblock Structural property-aware multilayer network embedding for latent factor analysis.
\newblock {\em Pattern Recognition}, 76:228--241, 2018.

\bibitem[\protect\citeauthoryear{Lu \bgroup \em et al.\egroup }{2019}]{lu2019fuzzy}
Jie Lu, Hua Zuo, and Guangquan Zhang.
\newblock Fuzzy multiple-source transfer learning.
\newblock {\em IEEE Transactions on Fuzzy Systems}, 28(12):3418--3431, 2019.

\bibitem[\protect\citeauthoryear{Nie \bgroup \em et al.\egroup }{2023}]{pachtst}
Yuqi Nie, Nam~H. Nguyen, Phanwadee Sinthong, and Jayant Kalagnanam.
\newblock A time series is worth 64 words: Long-term forecasting with transformers.
\newblock In {\em {ICLR}}, 2023.

\bibitem[\protect\citeauthoryear{Platt and others}{1999}]{platt1999probabilistic}
John Platt et~al.
\newblock Probabilistic outputs for support vector machines and comparisons to regularized likelihood methods.
\newblock {\em Advances in large margin classifiers}, 10(3):61--74, 1999.

\bibitem[\protect\citeauthoryear{Platt}{2000}]{2000Probabilistic}
John~C. Platt.
\newblock Probabilistic outputs for support vector machines and comparisons to regularized likelihood methods.
\newblock {\em Advances in Large Margin Classifiers}, 2000.

\bibitem[\protect\citeauthoryear{Qiu \bgroup \em et al.\egroup }{2024}]{li2024foundts}
Xiangfei Qiu, Jilin Hu, Lekui Zhou, Xingjian Wu, Junyang Du, Buang Zhang, Chenjuan Guo, Aoying Zhou, Christian~S. Jensen, Zhenli Sheng, and Bin Yang.
\newblock {TFB:} towards comprehensive and fair benchmarking of time series forecasting methods.
\newblock {\em {VLDB}}, 17(9):2363--2377, 2024.

\bibitem[\protect\citeauthoryear{Ronneberger \bgroup \em et al.\egroup }{2015}]{ronneberger2015u}
Olaf Ronneberger, Philipp Fischer, and Thomas Brox.
\newblock {U-Net}: Convolutional networks for biomedical image segmentation.
\newblock In {\em {MICCAI}}, volume 9351 of {\em Lecture Notes in Computer Science}, pages 234--241. Springer, 2015.

\bibitem[\protect\citeauthoryear{Sun \bgroup \em et al.\egroup }{2023}]{SunLL23}
Chunlin Sun, Linyu Liu, and Xiaocheng Li.
\newblock {P}redict-then-{C}alibrate: {A} {N}ew {P}erspective of {R}obust {C}ontextual {LP}.
\newblock In {\em NeurIPS}, 2023.

\bibitem[\protect\citeauthoryear{Tan \bgroup \em et al.\egroup }{2024}]{tan2024language}
Mingtian Tan, Mike~A Merrill, Vinayak Gupta, Tim Althoff, and Thomas Hartvigsen.
\newblock Are language models actually useful for time series forecasting?
\newblock In {\em {NeurIPS}}, 2024.

\bibitem[\protect\citeauthoryear{Wang \bgroup \em et al.\egroup }{2024}]{tslib}
Yuxuan Wang, Haixu Wu, Jiaxiang Dong, Yong Liu, Mingsheng Long, and Jianmin Wang.
\newblock Deep time series models: {A} comprehensive survey and benchmark.
\newblock {\em CoRR}, abs/2407.13278, 2024.

\bibitem[\protect\citeauthoryear{Wen \bgroup \em et al.\egroup }{2023}]{wen2023diffstg}
Haomin Wen, Youfang Lin, Yutong Xia, Huaiyu Wan, Qingsong Wen, Roger Zimmermann, and Yuxuan Liang.
\newblock Diffstg: Probabilistic spatio-temporal graph forecasting with denoising diffusion models.
\newblock In {\em Proceedings of the 31st ACM International Conference on Advances in Geographic Information Systems}, pages 1--12, 2023.

\bibitem[\protect\citeauthoryear{Wu \bgroup \em et al.\egroup }{2023}]{wu2022timesnet}
Haixu Wu, Tengge Hu, Yong Liu, Hang Zhou, Jianmin Wang, and Mingsheng Long.
\newblock Times{N}et: Temporal 2d-variation modeling for general time series analysis.
\newblock In {\em {ICLR}}. OpenReview.net, 2023.

\bibitem[\protect\citeauthoryear{Zeng \bgroup \em et al.\egroup }{2023}]{Zeng_Chen_Zhang_Xu_2023_dlnear}
Ailing Zeng, Muxi Chen, Lei Zhang, and Qiang Xu.
\newblock Are transformers effective for time series forecasting?
\newblock {\em Proceedings of the AAAI Conference on Artificial Intelligence}, 37(9):11121--11128, Jun. 2023.

\bibitem[\protect\citeauthoryear{Zhang \bgroup \em et al.\egroup }{2023}]{ZhangMDZ23}
Rongtao Zhang, Xueling Ma, Weiping Ding, and Jianming Zhan.
\newblock {MAP-FCRNN:} multi-step ahead prediction model using forecasting correction and {RNN} model with memory functions.
\newblock {\em Inf. Sci.}, 646:119382, 2023.

\bibitem[\protect\citeauthoryear{Zhang \bgroup \em et al.\egroup }{2024}]{zhang2024probts}
Jiawen Zhang, Xumeng Wen, Zhenwei Zhang, Shun Zheng, Jia Li, and Jiang Bian.
\newblock {ProbTS}: Benchmarking point and distributional forecasting across diverse prediction horizons.
\newblock In {\em {NeurIPS}}, 2024.

\bibitem[\protect\citeauthoryear{Zhong \bgroup \em et al.\egroup }{2025}]{zhong2025time}
Siru Zhong, Weilin Ruan, Ming Jin, Huan Li, Qingsong Wen, and Yuxuan Liang.
\newblock Time-vlm: Exploring multimodal vision-language models for augmented time series forecasting.
\newblock In {\em Proceedings of the 42nd International Conference on Machine Learning}, 2025.

\bibitem[\protect\citeauthoryear{Zhou \bgroup \em et al.\egroup }{2022}]{zhou2022fedformer}
Tian Zhou, Ziqing Ma, Qingsong Wen, Xue Wang, Liang Sun, and Rong Jin.
\newblock {FEDformer}: Frequency enhanced decomposed transformer for long-term series forecasting.
\newblock In {\em {ICML}}, volume 162 of {\em Proceedings of Machine Learning Research}, pages 27268--27286. {PMLR}, 2022.

\bibitem[\protect\citeauthoryear{Zhou \bgroup \em et al.\egroup }{2023a}]{zhou2023one}
Tian Zhou, Peisong Niu, Liang Sun, Rong Jin, et~al.
\newblock One fits all: Power general time series analysis by pretrained lm.
\newblock {\em Advances in neural information processing systems}, 36:43322--43355, 2023.

\bibitem[\protect\citeauthoryear{Zhou \bgroup \em et al.\egroup }{2023b}]{pTSE23}
Yunyi Zhou, Zhixuan Chu, Yijia Ruan, Ge~Jin, Yuchen Huang, and Sheng Li.
\newblock {pTSE}: {A} multi-model ensemble method for probabilistic time series forecasting.
\newblock In {\em {IJCAI}}, pages 4684--4692. ijcai.org, 2023.

\end{thebibliography}

\clearpage
\onecolumn

\renewcommand*\appendixpagename{\centering Appendix}
\begin{appendices}
\setcounter{figure}{0}
\setcounter{section}{0}
\renewcommand{\thefigure}{A\arabic{figure}}
\setcounter{table}{0}
\renewcommand{\thetable}{A\arabic{table}}
\section{Analysis of Efficient Parallel Training in SoP}\label{sub:analysis}
In practice, the computational cost and training time of the non-collective, variable-wise SoP strategy remain efficient compared to the collective calibration. As each of the \(N\) Plugs calibrates its corresponding target variable $i$ independently, the total training loss \( \mathcal{L}^{train} \) is defined as the sum of the individual losses:
\[
\mathcal{L}^{train} = \sum_{i=1}^{N} \mathcal{L}_i^{train}
\]
Accordingly, the gradient of the total loss during backpropagation is:
\[
\nabla \mathcal{L}^{train}(\theta) = \sum_{i=1}^{N} \nabla \mathcal{L}_i^{train}(\theta_i)
\]
This formulation inherently computes the gradient for each \(\text{Plug}_i\) to its variable $i$, independently, i.e., \(\nabla \mathcal{L}_i^{train}(\theta_i)\). Such independent gradient computation not only facilitates parallel training of all Plugs—seamlessly managed by PyTorch for efficiency—but also ensures that the optimization of one target variable does not interfere with others, effectively avoiding conflicting gradients.

Owing to this parallelism, the overall training duration of variable-wise SoP is determined by the last Plug to meet its individual early stopping condition, introducing only a minor delay relative to collective calibration—yet it consistently achieving superior performance.

\section{Experiments on Time Series Forecasting}\label{sub:deatil_series}
We conduct extensive experiments on benchmark time series datasets, including ETTh1, ETTh2, ECL, Exchange, Weather, and Solar-Energy. Details of these datasets are introduced below.
\subsection{Time series dataset}\label{Time series dataset}
\begin{itemize}
    \item ETTh1/ETTh2: these two datasets are hourly collected from two separate counties in China for electricity transformer temperature, where each contains the oil temperature and 6 power load indicators from July 2016 to July 2018.
    \item Electricity: the electricity dataset contains the hourly electricity consumption of 321 customers from 2012 to 2014.
    \item Traffic: the traffic dataset comprises hourly road occupancy rate data collected over two years by various sensors installed on freeways in the San Francisco Bay Area. This dataset is provided by the California Department of Transportation.
    \item Weather: the weather dataset consists of 21 meteorological indicators, including temperature, humidity, and others, recorded at 10-minute intervals throughout the entire year of 2020.
    \item Exchange: the exchange dataset contains panel data of daily exchange rates from 8 countries, spanning the period from 1990 to 2016.
    \item Solar-energy: the solar-energy dataset captures the solar power production of 137 photovoltaic (PV) plants in 2006, sampled at 10-minute intervals.
\end{itemize}
 
\subsection{Inference models} \label{baseline_models}
Below is a brief overview of the inference models selected for our study:
\begin{itemize}
    \item SOFTS: it employs the centralized strategy to learn the interactions between different time series.
    \item iTransformer: it utilizes an inverted Transformer architecture to support multi-variable and multi-step predictions, which leverages the attention mechanism to capture correlations among variables.
    \item FEDformer: it introduces a sparse representation in the frequency domain and incorporates frequency-enhanced blocks to effectively capture cross-time dependencies.
    \item TimesNet: it transforms time series data into a 2D space and employs CNNs to capture multi-scale temporal patterns.
    \item DLinear: it leverages seasonal-trend decomposition and the linear model to extract temporal dependencies.
    \item PatchTST: it utilizes Transformer encoders as its backbone and introduces patching and channel independence techniques to enhance prediction performance.
\end{itemize}

\subsection{Hyperparameter settings}\label{timeHyperparameter settings}
In each Plug, the hidden dimension of the MLP is set to $d = 256$. The Adam optimizer is employed with an initial learning rate of 1e-4, and the L2 loss is used for optimizing the Plug module. The early-stopping patience is set to 5 epochs. The past time step $T$ is fixed at 96. Prediction horizons across all datasets follow the protocols established in prior studies, with $S \in {96, 192, 336, 720}$. All experiments are conducted on a single NVIDIA GeForce RTX 4090 GPU with 24 GB VRAM.

\subsection{Performance of SoP at different prediction horizons}\label{different prediction horizons}
To provide more detailed experimental results, we also evaluated the Weather and Traffic datasets, demonstrating that a Plug with consistent parameters across different prediction horizons achieves significant performance improvements. 

\begin{figure*}[ht]
  \centering
  \includegraphics[width=0.65\linewidth]{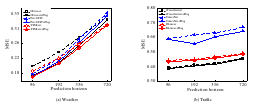}
  \caption{Forecasting performance with the fixed past time step T = 96, adding SoP at different prediction horizons \({S = \{96, 192, 336, 720\}}\), (a-b) display the test results (MSE) for the Weather and Traffic datasets, respectively.}
  \label{fig：weather-traffic}
\end{figure*}


\subsection{Performance of SoP compared to LLM methods}\label{sub:llm}
The experimental results in Table \ref{tab:my-table3} demonstrate that the iTransformer model enhanced with variable-wise SoP achieves competitive performance compared to state-of-the-art (SOTA) LLM-based methods, including CALF~\cite{liu2024calf}, TimeLLM~\cite{timellm}, and GPT4TS~\cite{zhou2023one}, for time series forecasting. Notably, iTransformer+SoP outperforms all LLM-based methods on the Traffic and Weather datasets, achieving the lowest MSE and MAE. On other datasets, such as ETTh1 and ETTh2, while CALF achieves slightly better results, iTransformer+SoP consistently reduces the performance gap. Previous studies have highlighted that optimizing existing deep forecasting models typically requires extensive architecture search and hyperparameter tuning~\cite{pTSE23}, which has driven the adoption of LLM-based methods for time series prediction. However, these methods come with significant training and inference overheads. In contrast, SoP leverages the pre-trained outputs of the Socket without requiring complex architectural modifications or hyperparameter adjustments, achieving performance improvements with minimal resource costs. These findings highlight the potential of enhancing SOTA backbones with SoP as a lightweight and competitive alternative to LLMs for time series forecasting tasks.


\begin{table}[H]
\begin{center}
\resizebox{0.58\columnwidth}{!}{%
\begin{tabular}{ccccccccc}
\hline
Models &
  \multicolumn{2}{c}{iTransformer+Plug} &
  \multicolumn{2}{c}{CALF} &
  \multicolumn{2}{c}{TimeLLM} &
  \multicolumn{2}{c}{GPT4TS} \\ \hline
Metric &
  MSE &
  MAE &
  MSE &
  MAE &
  MSE &
  MAE &
  MSE &
  MAE \\ \hline
ETTh1 &
  0.450 &
  0.447 &
  {\color[HTML]{FF0000} \textbf{0.432}} &
  {\color[HTML]{FF0000} \textbf{0.428}} &
  0.460 &
  0.449 &
  \textbf{0.447} &
  \textbf{0.436} \\ \hline
ETTh2 &
  \textbf{0.381} &
  \textbf{0.404} &
  {\color[HTML]{FF0000} \textbf{0.349}} &
  {\color[HTML]{FF0000} \textbf{0.382}} &
  0.389 &
  0.408 &
  \textbf{0.381} &
  0.408 \\ \hline
Traffic &
  {\color[HTML]{FF0000} \textbf{0.415}} &
  {\color[HTML]{FF0000} \textbf{0.279}} &
  \textbf{0.439} &
  \textbf{0.281} &
  0.541 &
  0.358 &
  0.488 &
  0.317 \\ \hline
Weather &
  {\color[HTML]{FF0000} \textbf{0.249}} &
  {\color[HTML]{FF0000} \textbf{0.274}} &
  \textbf{0.250} &
  {\color[HTML]{FF0000} \textbf{0.274}} &
  0.274 &
  0.290 &
  0.264 &
  0.284 \\ \hline
\end{tabular}
}
\caption{Compared to the existing LLM methods for time series forecasting.}
\label{tab:my-table3}
\end{center}
\end{table}

\subsection{Equitable settings for parameter size of SoP with different plug counts}\label{sub:count}
In this section, we detail the parameter settings for the Plug modules to ensure fair fitting capacity across different configurations. For instance, in the Weather dataset, which contains 21 variables, the number of Plugs corresponds to the chosen calibration strategy. As described in \textit{Remark 1}, setting the Plug count equal to the total number of variables results in the variable-wise SoP approach, where Plug\(_1\) to Plug\(_{21}\) are assigned to variables Var\(_1\) through Var\(_{21}\), respectively. This configuration, illustrated in the upper part of Figure~\ref{fig:group1}, utilizes 21 independent Plugs. Conversely, as explained in \textit{Remark 2}, reducing the Plug count to one corresponds to the classical collective calibration strategy, where a single Plug is responsible for all variables. The parameter size of this single Plug {is set} 21 times that of an individual Plug in the variable-wise SoP {to ensure a roughly fair fitting capability}. For intermediate configurations, such as Plug counts of 3 or 7, the Plugs are applied to grouped variables. For example, when the Plug count is set to 3, each Plug is responsible for predicting seven variables: Plug\(_A\) handles Var\(_1\) through Var\(_7\), Plug\(_B\) handles Var\(_8\) through Var\(_{14}\), and Plug\(_C\) handles Var\(_{15}\) through Var\(_{21}\). This configuration, depicted in the lower part of Figure~\ref{fig:group1}, involves three Plugs, each with a parameter size 7 times that of a single Plug in the variable-wise SoP (e.g., the parameter count of Plug\(_A\) equals the combined parameter size of Plug\(_1\) through Plug\(_7\)).

\begin{figure*}[hb] \centering
\begin{minipage}[t]{0.48\linewidth}
    \includegraphics[width=\textwidth,height=0.65\textwidth]{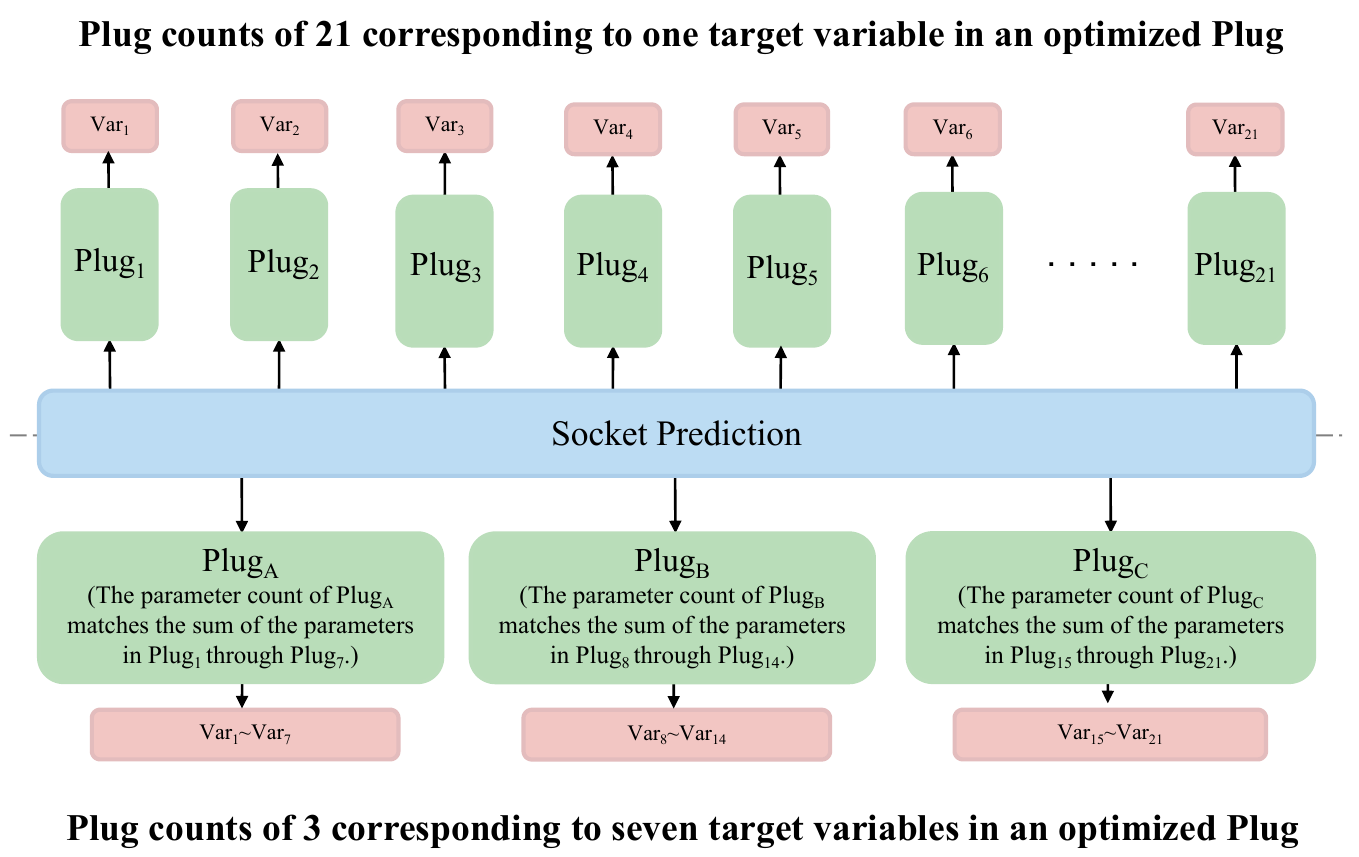}
    \caption{Illustration of the experimental setup with different Plug count.} \label{fig:group1}
\end{minipage}\hfill
\begin{minipage}[t]{0.48\linewidth}
    \includegraphics[width=\textwidth,height=0.65\textwidth]{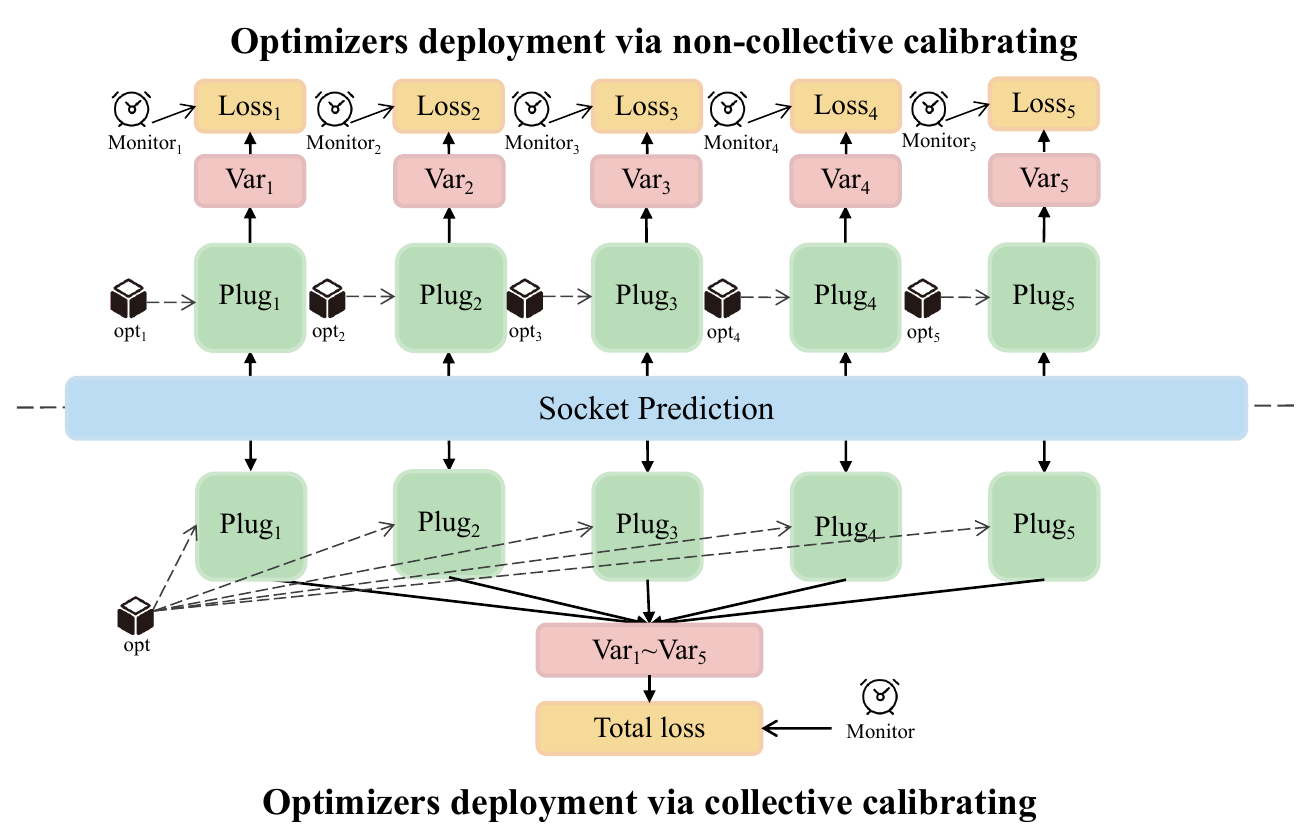}
    \caption{Illustration of the ablation study for non-collective and collective calibrating.} \label{fig：group3}
\end{minipage}
\end{figure*}

\subsection{Ablation study of non-collective versus collective calibrating}\label{sub:collective}
Figure \ref{fig：group3} illustrates the deployment of optimizers in the ablation study for both non-collective and collective calibration. In the non-collective calibration, each Plug$_i$ is paired with a dedicated optimizer, opt$_i$, and an early-stopping monitor, Monitor$_i$, for the validation loss of variable Var$_i$. In contrast, for collective calibration, all Plugs share a single optimizer and monitor, which are applied to the total validation loss.

Figure \ref{fig:early} illustrates the difference in the number of training epochs required to trigger early stopping when deploying non-collective and collective optimizers. The experiments were conducted on the Exchange dataset, which consists of eight variables. We compare the epochs required to trigger early stopping in both non-collective and collective calibrating setups. The experimental results are shown in Figure \ref{fig:early}, where panels ${(a \sim d)}$ use DLinear as the Socket, and panels ${(e \sim h)}$ use iTransformer as the Socket. In both cases, it can be observed that, with non-collective calibrating, some variables trigger early stopping in very few epochs, while others require significantly more epochs to fine-tune the parameters. This suggests that during the execution of SoP, not all Plugs concurrently converge to optimized parameters. In contrast, with collective calibrating, early stopping is triggered more quickly, but this may not yield the optimal validation loss for each variable. Specifically, for the same prediction horizon $S$, the DLinear and iTransformer models exhibit distinct early stopping behaviors. For example, when the prediction horizon $S = 96$, Var$_5$ and Var$_6$ of the DLinear model are the first to satisfy the early stopping criteria, whereas Var$_1$ of the iTransformer model stops first. Meanwhile, Var$_4$ in both models exhibits delayed convergence. The longer convergence time of this Plug may indicate that Var$_4$ is more difficult to learn compared to the other variables.

\begin{figure}[ht]
  \centering
  \includegraphics[width=\linewidth]{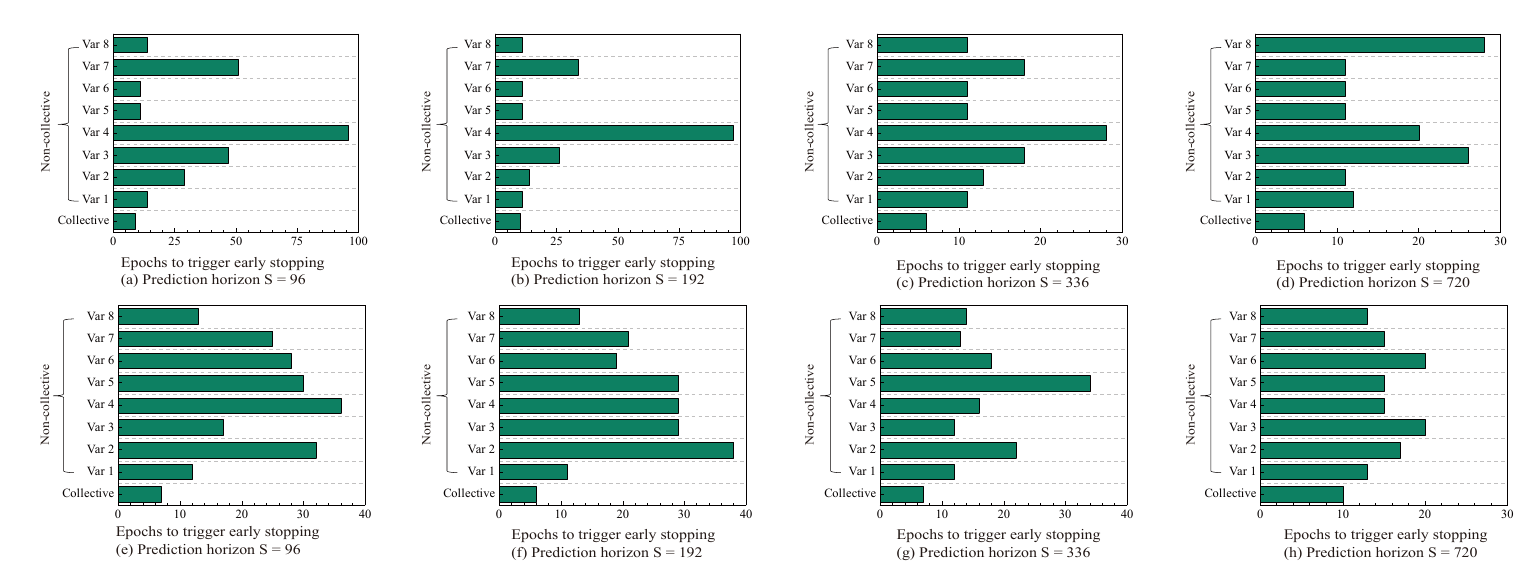}
  \caption{Non-collective and collective calibrating are employed on DLinear (a-d) and iTransformer (e-h) models to demonstrate the epochs at which early stopping is triggered.}
  \label{fig:early}
\end{figure}





\subsection{Robustness of SoP}\label{sub:robust}
The robustness of SoP was evaluated through five repeated experiments, each initialized with a different random seed. The iTransformer model served as the Socket for this demonstration. The experimental results, reported as "mean ± std" in Table \ref{tab:lb}, demonstrate a small standard deviation, which indicates the considerable robustness of SoP.

\begin{table}[H]
\centering
\resizebox{\linewidth}{!}{
\begin{tabular}{ccccccccc}
\hline
Dataset &
  \multicolumn{2}{c}{ETTh1} &
  \multicolumn{2}{c}{ETTh2} &
  \multicolumn{2}{c}{Weather} &
  \multicolumn{2}{c}{Exchange} \\ \cline{2-9} 
Horizon &
  MSE &
  MAE &
  MSE &
  MAE &
  MSE &
  MAE &
  MSE &
  MAE \\ \hline
96 &
  0.3810 ${\pm}$ 0.001 &
  0.4026 ${\pm}$ 0.0006 &
  0.2986 ${\pm}$ 0.001 &
  0.3456 ${\pm}$ 0.0005 &
  0.1625 ${\pm}$ 0.0001 &
  0.2053 ${\pm}$ 0.0001 &
 0.0881 ${\pm}$ 0.002 &
  0.2124${\pm}$0.003 \\ \hline
192 &
  0.4369 ${\pm}$ 0.001 &
  0.4315 ${\pm}$ 0.0005 &
  0.3786 ${\pm}$ 0.002 &
  0.3945 ${\pm}$ 0.004 &
  0.2115 ${\pm}$ 0.001 &
  0.2504 ${\pm}$ 0.001 &
  0.1712 ${\pm}$ 0.004 &
  0.3045 ${\pm}$ 0.003 \\ \hline
336 &
  0.4844 ${\pm}$ 0.002 &
  0.4572 ${\pm}$ 0.0005 &
  0.4305 ${\pm}$ 0.004 &
  0.4313 ${\pm}$ 0.001 &
  0.2690 ${\pm}$ 0.0005 &
  0.2924 ${\pm}$ 0.0002 &
  0.2735 ${\pm}$ 0.001 &
  0.3998 ${\pm}$ 0.003 \\ \hline
720 &
  0.4926 ${\pm}$ 0.003 &
  0.4783 ${\pm}$ 0.0005 &
  0.4326 ${\pm}$ 0.001 &
  0.4457 ${\pm}$ 0.0002 &
  0.3524 ${\pm}$ 0.0002 &
  0.3476 ${\pm}$ 0.001 &
  0.6749 ${\pm}$ 0.03 &
  0.6487 ${\pm}$ 0.005 \\ \hline
\end{tabular}
}
\caption{Robustness of SoP. Results are reported as "mean ± std" over five repeated experiments with different random seeds.}
\label{tab:lb}
\end{table}

\subsection{Transferability of the trained Plug}\label{sub:transfer}
Considering that the Socket itself exhibits predictive capabilities, we aimed to investigate whether transferring a trained Plug from one Socket (e.g., DLinear) to another could similarly improve performance. To explore this, we conducted experiments on the Exchange dataset. The results, presented in Table~\ref{tab:pinjie}, provide initial evidence supporting the transferability of the Plug across different Sockets. Specifically, when the optimal Plugs are transferred to other Sockets, they continue to improve predictive performance across nearly all horizons. This demonstrates the existence of a transferable calibrating strategy that can rapidly yield significant results on different Sockets. However, experiments conducted on other datasets also revealed that the Plug, which performs optimally (i.e., achieving the lowest MSE and MAE following SoP), may experience performance degradation in certain cases. Nevertheless, this still highlights the Plug's favorable cross-Socket capability. One potential benefit is that transferring a trained specific Plug can significantly reduce the computational resources required to re-train the entire model from scratch, which is particularly valuable in real-world applications where resources are limited.



\begin{table}[H]
\centering
\resizebox{0.98\columnwidth}{!}{%
\begin{tabular}{cccccccccccccc}
\hline
\multicolumn{2}{c}{Model of Socket} &
  \multicolumn{2}{c}{SOFTS} &
  \multicolumn{2}{c}{iTransformer} &
  \multicolumn{2}{c}{PatchTST} &
  \multicolumn{2}{c}{TimesNet} &
  \multicolumn{2}{c}{TSMixer} &
  \multicolumn{2}{c}{FEDformer} \\ \hline
\multicolumn{1}{c}{Horizon} &
  Metric &
  MSE &
  MAE &
  MSE &
  MAE &
  MSE &
  MAE &
  MSE &
  MAE &
  MSE &
  MAE &
  MSE &
  MAE \\ \hline
\multicolumn{1}{c}{} &
  Socket &
  0.097 &
  0.220 &
  0.089 &
\textbf{0.210} &
  0.086 &
  0.203 &
  0.107 &
  0.237 &
  0.194 &
  0.353 &
  0.161 &
  0.290 \\
\multicolumn{1}{c}{\multirow{-2}{*}{S=96}} &
  Socket+DLinear's Plug &
  \textbf{0.096} &
  \textbf{0.217} &
  \textbf{0.085} &
  \textbf{0.210} &
  \textbf{0.083} &
  \textbf{0.202} &
  \textbf{0.104} &
  \textbf{0.233} &
  \textbf{0.185} &
  \textbf{0.348} &
  \textbf{0.160} &
  \textbf{0.287} \\ \hline
\multicolumn{1}{c}{} &
  Socket &
  0.200 &
  0.323 &
  0.186 &
  0.309 &
  0.178 &
  0.300 &
  0.214 &
  0.336 &
  0.361 &
  0.484 &
  0.290 &
  0.393 \\
\multicolumn{1}{c}{\multirow{-2}{*}{S=192}} &
  Socket+DLinear's Plug &
  \textbf{0.197} &
  \textbf{0.321} &
  \textbf{0.184} &
  \textbf{0.307} &
  \textbf{0.173} &
  \textbf{0.296} &
  \textbf{0.206} &
  \textbf{0.330} &
  \textbf{0.337} &
  \textbf{0.470} &
  \textbf{0.282} &
  \textbf{0.389} \\ \hline
\multicolumn{1}{c}{} &
  Socket &
  0.325 &
  0.415 &
  0.357 &
  0.433 &
  0.357 &
  0.434 &
  0.378 &
  0.448 &
  0.544 &
  0.603 &
  0.456 &
  0.522 \\
\multicolumn{1}{c}{\multirow{-2}{*}{S=336}} &
  Socket+DLinear's Plug &
  \textbf{0.316} &
  \textbf{0.409} &
  \textbf{0.339} &
  \textbf{0.424} &
  \textbf{0.341} &
  \textbf{0.425} &
  \textbf{0.370} &
  \textbf{0.445} &
  \textbf{0.511} &
  \textbf{0.587} &
  \textbf{0.440} &
  \textbf{0.487} \\ \hline
\multicolumn{1}{c}{} &
  Socket &
  1.028 &
  \textbf{0.750} &
  0.897 &
  0.720 &
  0.940 &
  0.732 &
  0.989 &
  0.760 &
  0.674 &
  \textbf{0.680} &
  1.164 &
  0.827 \\
\multicolumn{1}{c}{\multirow{-2}{*}{S=720}} &
  Socket+DLinear's Plug &
  \textbf{1.015} &
  0.762 &
  \textbf{0.888} &
  \textbf{0.718} &
  \textbf{0.916} &
  \textbf{0.727} &
  \textbf{0.986} &
  \textbf{0.759} &
  \textbf{0.661} &
  0.682 &
  \textbf{1.138} &
  \textbf{0.825} \\ \hline
\multicolumn{2}{c}{Socket (Avg)} &
  0.413 &
  {\color[HTML]{FE0000} \textbf{0.427}} &
  0.382 &
  0.418 &
  0.390 &
  0.417 &
  0.422 &
  0.445 &
  0.443 &
  0.530 &
  0.518 &
  0.508 \\
\multicolumn{2}{c}{Socket+DLinear's Plug (Avg)} &
  {\color[HTML]{FE0000} \textbf{0.406}} &
  {\color[HTML]{FE0000} \textbf{0.427}} &
  {\color[HTML]{FE0000} \textbf{0.375}} &
  {\color[HTML]{FE0000} \textbf{0.415}} &
  {\color[HTML]{FE0000} \textbf{0.379}} &
  {\color[HTML]{FE0000} \textbf{0.412}} &
  {\color[HTML]{FE0000} \textbf{0.417}} &
  {\color[HTML]{FE0000} \textbf{0.442}} &
  {\color[HTML]{FE0000} \textbf{0.424}} &
  {\color[HTML]{FE0000} \textbf{0.522}} &
  {\color[HTML]{FE0000} \textbf{0.408}} &
  {\color[HTML]{FE0000} \textbf{0.471}} \\ \hline
\end{tabular}
}
\caption{Direct transferring the trained Plug of DLinear to other Socket models.}
\label{tab:pinjie}
\end{table}

\section{Experiments on Spatio-temporal Forecasting}\label{Spatio-temporal dataset}

\subsection{Spatio-temporal dataset}
The meteorological ERA5 dataset~\cite{tianchi21} covers a period of 10 years, from 2007 to 2016, with a spatial resolution of 0.25°, spanning a geographic coverage area of \(N10 \sim N50, E100 \sim E140\), and a temporal resolution of 6 hours. It includes five surface-level measurements as target variables: the temperature at 2 meters above ground (T2M), the u-component (U10) and v-component (V10) of wind speed at 10 meters, mean sea level pressure (MSL), and total precipitation (TP). The dataset is split into three subsets: the samples of the first eight years (2007–2014) for training, the samples of the year 2015 for validation, and the samples from the year 2016 for testing.

\subsection{Inference models and hyperparameter settings}\label{st hyperparameter settings}
The input channels of the Unet model~\cite{ronneberger2015u} are set to 10 (\(T \times N = 2 \times 5\)),  and the output channels are set to 100 (\(S \times N = 20 \times 5\)). We used Adam with an initial learning rate of 1e-3 and L2 loss for Unet optimization, and Adam with an initial learning rate of 1e-4 and L2 loss for Plug calibration.  Early stopping is applied with a patience of 10 epochs for monitoring.



\subsection{Case study of meteorological forecasting}\label{sub:variables}
We conduct a case study on the meteorological dataset using Unet with SoP. 
Figure \ref{fig:5201} reports the prediction performance of five surface variables 
over 5 days. We can observe that the prediction errors are significantly lower across the majority of time steps compared to those of the original Unet model. Although the MSE of  Unet+Plug increases at certain horizons, the performance remains comparable to that 
of the original model. Furthermore, the errors at the final step for all five variables are substantially reduced.



\begin{figure*}[ht]
  \centering
  \includegraphics[width=\linewidth]{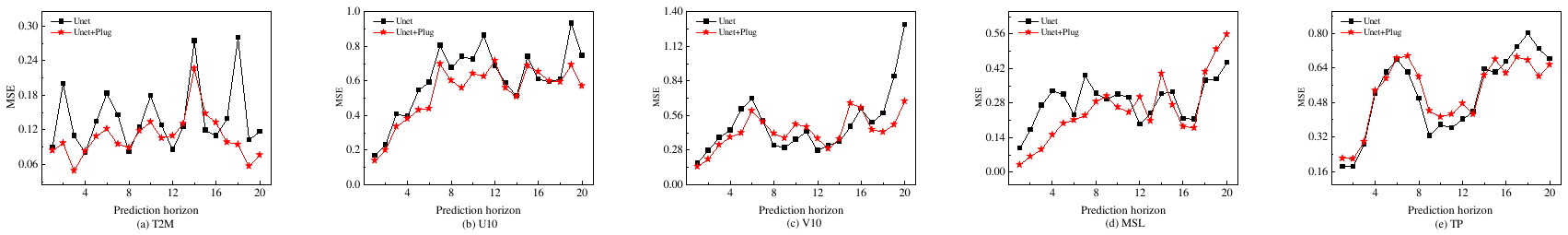}
  \caption{A test example to illustrate the prediction performance of the five surface variables over 20 subsequent prediction horizons with the integration of step-wise SoP. (a-e) depict the MSE variations for T2M, U10, V10, MSL, and TP, respectively.}
  \label{fig:5201}
\end{figure*}

\end{appendices}

\end{document}